\documentclass[conference]{IEEEtran}
\usepackage{graphicx} 
\usepackage{cite}
\usepackage{booktabs}
\usepackage{multirow}
\usepackage{gensymb}
\usepackage{amsmath}
\usepackage{amssymb}
\usepackage{rotating}
\usepackage{adjustbox}
\usepackage{url}
\usepackage{xcolor}

\title{ChemDIRT: A Diversified Instruction, Representation, and Task Benchmark for Robust Chemistry-LLM Evaluation}

\author{\IEEEauthorblockN{Eric Inae, Tim Gunn, Chris Bond, and Meng Jiang}
\IEEEauthorblockA{University of Notre Dame\\
Notre Dame, Indiana, USA\\
\{einae, tgunn2, cbond, mjiang2\}@nd.edu}}

\begin{document}

\maketitle

\begin{abstract}
The rapid advancement of large language models (LLMs) has led to increasing interest in their application to scientific domains such as chemistry. However, existing chemistry benchmarks often provide only a narrow view of model capability, focusing on limited task sets while overlooking robustness to variations in problem formulation and chemical representation. As a result, reported performance may overestimate a model’s true ability to reason consistently across realistic settings. To address this challenge, we introduce ChemDIRT (Diversified Instruction, Representation, and Task Benchmark), a comprehensive evaluation framework designed to assess the robustness of chemical reasoning in LLMs. ChemDIRT systematically measures model performance across variations in instructions and molecular representations while spanning eight categories of chemistry tasks. By evaluating both accuracy and consistency under these controlled perturbations, ChemDIRT provides a more reliable assessment of model reasoning capabilities than conventional single-format benchmarks. We benchmark a diverse set of open- and closed-source LLMs, revealing substantial prompt sensitivity, representation dependence, and uneven performance across task families.
\end{abstract}


\section{Introduction}

The rapid adoption of large language models (LLMs) as agents for general reasoning and question-answering has increasingly led to their application in specialized scientific domains \cite{xiang2025survey, ren2025towards}. For chemistry, LLMs have been leveraged for tasks such as molecular design and discovery \cite{narayanan2025training, bran2023chemcrow}, quantitative analysis \cite{zhou2025reso}, and predictive chemistry \cite{lin2025enhancing, zhang2024chemllm}. As these models become integrated into scientific workflows, rigorous evaluation becomes increasingly important. In particular, reliable deployment requires not only strong performance on benchmark tasks, but also robustness to variations in task formulation, linguistic expression, and molecular representation. An evaluation framework that fails to account for these factors risks providing an incomplete or misleading assessment of a model's underlying chemical reasoning capabilities.

Existing chemistry benchmarks typically focus on a narrow range of reasoning tasks. In this work, we create a taxonomy that organizes these tasks by their primary reasoning requirements: description, classification, design, analytical, regression, computation, mechanistic, and relational. For example, design tasks generate molecular structures, regression tasks predict real-valued properties, and computation tasks require symbolic or numerical reasoning. Existing benchmarks such as ChemLLMBench \cite{guo2023can}, ChemBench4K \cite{zhang2024chemllm}, and Mol-Instructions \cite{fang2023mol} cover only subsets of these categories. Strong performance on any one benchmark may therefore not reflect robust reasoning across the range of tasks in chemistry.

\begin{figure}[t]
\includegraphics[width=\columnwidth]{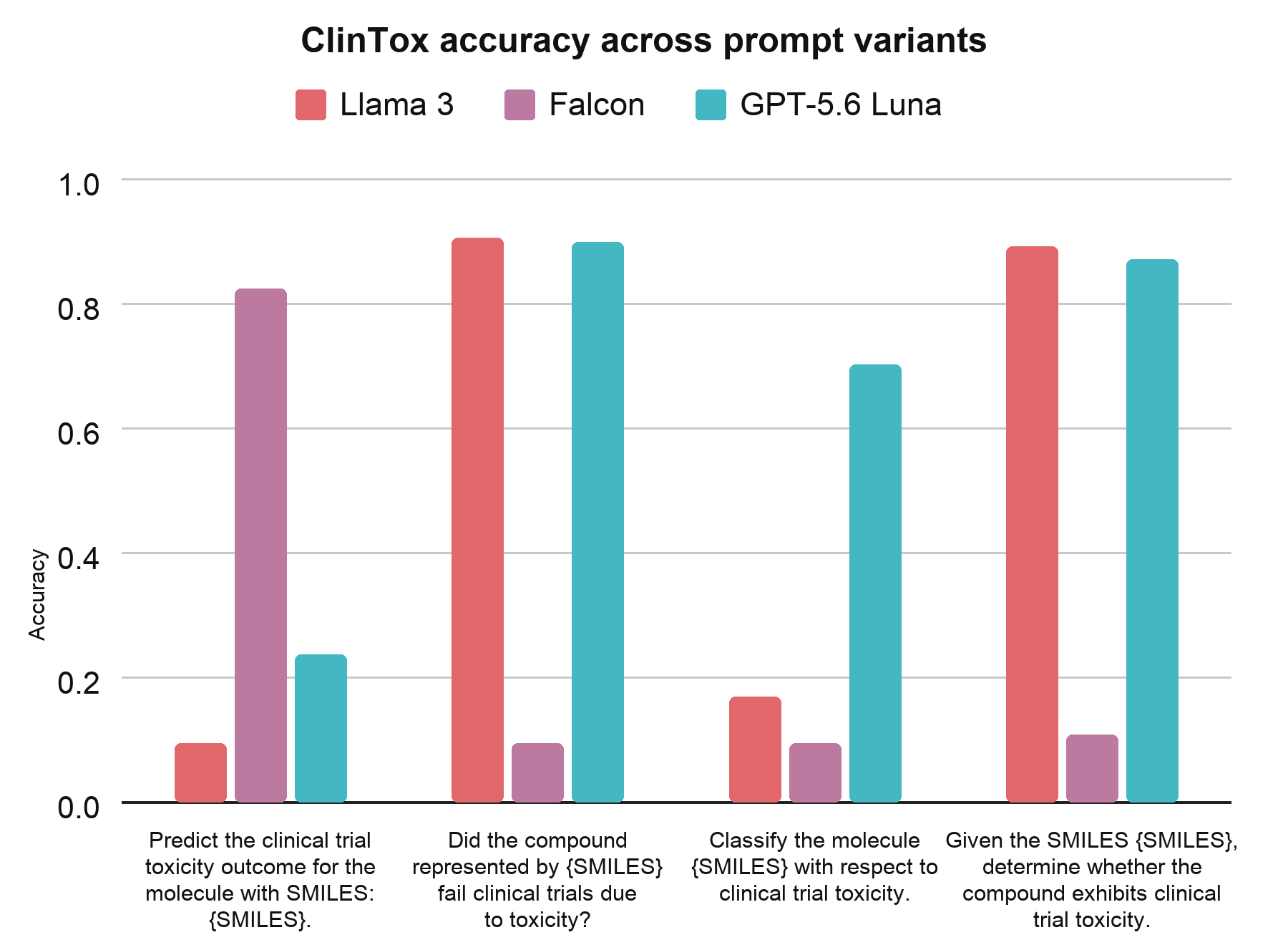}
\caption{ClinTox classification accuracy across four semantically equivalent prompt variants. Despite asking the same question, changes in phrasing produce substantial performance shifts across all three models.}
\label{fig:prompt_difference}
\vspace{-12pt}
\end{figure}

Beyond task coverage, current chemistry benchmarks rarely evaluate robustness to instruction variation. Previous works have found that LLMs are highly sensitive to choice of wording and syntactic ordering, despite the semantic content being equivalent \cite{weber2023mind, sclar2023quantifying, mizrahi2024state, polo2024efficient}. 
Nevertheless, many chemistry benchmarks employ a single fixed template for each task, implicitly conflating task performance with prompt-specific behavior. Figure \ref{fig:prompt_difference} illustrates this phenomenon for ClinTox classification, where semantically equivalent prompts yield substantially different accuracies. Consequently, benchmark results may reflect sensitivity to prompt formulation rather than genuine differences in chemical reasoning ability.
Several studies call for multi-prompt evaluation to address this sensitivity \cite{polo2024efficient, mizrahi2024state}. However, this practice remains largely absent from chemistry-specific benchmarks.

A related but often overlooked source of evaluation instability arises from molecular representation. While SMILES strings are the default choice for representing molecules, recent works have shown that alternative notations, including SELFIES, InChI, and IUPAC, can elicit different behavior from LLMs. Although a model should draw the same conclusion from the SMILES and IUPAC forms of a molecule, models exhibit low consistency across representations in practice \cite{yan2025inconsistency}.
Furthermore, alternative representations such as InChI and IUPAC have been reported to outperform SMILES on certain prediction tasks due to differences in tokenization characteristics and representation prevalence within pretraining corpora \cite{baker2025molecular}. Because most existing benchmarks evaluate models using only a single representation format, they are unable to determine whether observed performance reflects robust chemical understanding or dependence on a particular encoding. Evaluating representational consistency is therefore essential for obtaining a reliable assessment of model capability.

To address these limitations, we introduce ChemDIRT (\textbf{D}iversified \textbf{I}nstruction, \textbf{R}epresentation, and \textbf{T}ask), a robustness-oriented benchmark containing 33 task--representation settings and 4,484 released examples across eight reasoning categories, evaluated with 23 open- and closed-source models. ChemDIRT treats instruction and representation diversity as a benchmark design requirement, as seemingly superficial choices can materially change measured performance.

Our contributions are fourfold: (1) a taxonomy spanning eight categories of chemical reasoning, (2) a benchmark that jointly varies tasks, instructions, molecular representations, and input modalities (text and images), (3) a broad evaluation showing substantial task- and prompt-dependent inconsistency, and (4) a public dataset and evaluation codebase supporting reproducible scientific-model comparison. Code and data are available on GitHub (\url{https://github.com/einae-nd/ChemDIRT}) and Hugging Face (\url{https://huggingface.co/datasets/Einae/ChemDIRT}).

\section{Related Work}

The rapid development of chemistry-focused language models has motivated the creation of benchmarks that evaluate chemical understanding beyond general linguistic competence. Existing evaluations aim to probe how well models capture chemistry-specific knowledge grounded in molecular structure, symbolic representations, and experimental observations, rather than surface-level text correlations. Prior work spans a broad range of problem settings, including molecular recognition, qualitative inference, numerical prediction, algorithmic calculation, and reasoning about chemical reactions and processes \cite{ChemistryQA, mirza2025framework, runcie2025assessing, li2025beyond, hao2025can, yang2025r1}. Together, these benchmarks reflect a shared goal of assessing whether models can meaningfully reason about chemical entities and transformations.

A large body of work focuses on recognizing molecular features and making qualitative inferences from structured representations such as SMILES strings, molecular graphs, or textual descriptions. Common evaluations include identifying atoms, functional groups, rings, and structural motifs, as well as predicting categorical properties such as toxicity, bioactivity, or permeability \cite{krallinger2015chemdner, wu2018moleculenet, guo2023can, yu2024llasmol, zhong2025benchmarking, zhu2025moltextnet}. Datasets from MoleculeNet, including HIV, BBBP, and Tox21, have served as foundational benchmarks for predictive chemistry and have been adapted for instruction-following and LLM-based settings \cite{wu2018moleculenet, guo2023can, yu2024llasmol, zhong2025benchmarking, zhu2025moltextnet}. Closely related are molecular generation and translation tasks, where models produce molecular structures, descriptions, or alternative representations under specified constraints, as explored in ChemLLMBench, Mol-Instructions, MolQA, and related works \cite{guo2023can, fang2023mol, liu2024multimodal, degtyarenko2007chebi}.

Beyond surface-level recognition, several benchmarks target chemistry question answering and analysis that require integrating conceptual knowledge across subfields. General-purpose reasoning suites such as MMLU and MMLU-Pro include chemistry subsets spanning undergraduate- and graduate-level topics \cite{hendrycks2020measuring, wang2024mmlu}, while domain-specific datasets address literature-based question answering, hypothesis generation, and molecular property analysis \cite{jin2019pubmedqa, wellawatte2025chemlit, yang2024moose, ma2025general}. These tasks typically require recalling and applying chemical principles to new contexts, rather than direct extraction from a single representation, and are often used to probe broader chemical reasoning capabilities in language models \cite{mirza2025framework, rein2024gpqa, du2025supergpqa}.

Another major line of work focuses on numeric and process-oriented reasoning in chemistry. Regression benchmarks evaluate the prediction of continuous-valued properties derived from experiments or simulations, including solubility, lipophilicity, hydration free energy, and quantum mechanical quantities, using datasets such as ESOL, Lipo, FreeSolv, QM9, and Open Catalyst \cite{ruddigkeit2012enumeration, mobley2014freesolv, wu2018moleculenet, tran2023open}. Complementary computational benchmarks assess chemistry-grounded calculations, including stoichiometry, unit conversions, yield estimation, and spectroscopy-related analysis \cite{ChemistryQA, valenti2024chemalgebra, runcie2025assessing}. Mechanistic reasoning tasks further extend this scope by focusing on chemical reactions and transformations, drawing on large-scale reaction corpora such as USPTO and the Open Reaction Database to evaluate forward reaction prediction, retrosynthesis planning, and condition selection \cite{Lowe2017, kearnes2021open, schneider2016s, coley2017prediction, chen2020retro, wigh2024orderly}.

An important yet relatively unexplored evaluation task is comparative and relational reasoning between molecules. While some recent benchmarks incorporate relational tasks such as ranking the best reagents for a reaction and selecting the molecule most similar to a given description \cite{guo2023can, zhang2024chemllm}, they remain relatively underrepresented in existing evaluations.

While existing benchmarks probe many aspects of chemical knowledge, they are typically designed in isolation and emphasize narrow task families or representations. As a result, current evaluations provide an incomplete and often inconsistent picture of a model’s overall chemical reasoning ability, motivating the construction of a benchmark that jointly spans diverse chemistry tasks under a common evaluation setting.

\begin{figure}[t]
\centering
\includegraphics[width=\columnwidth]{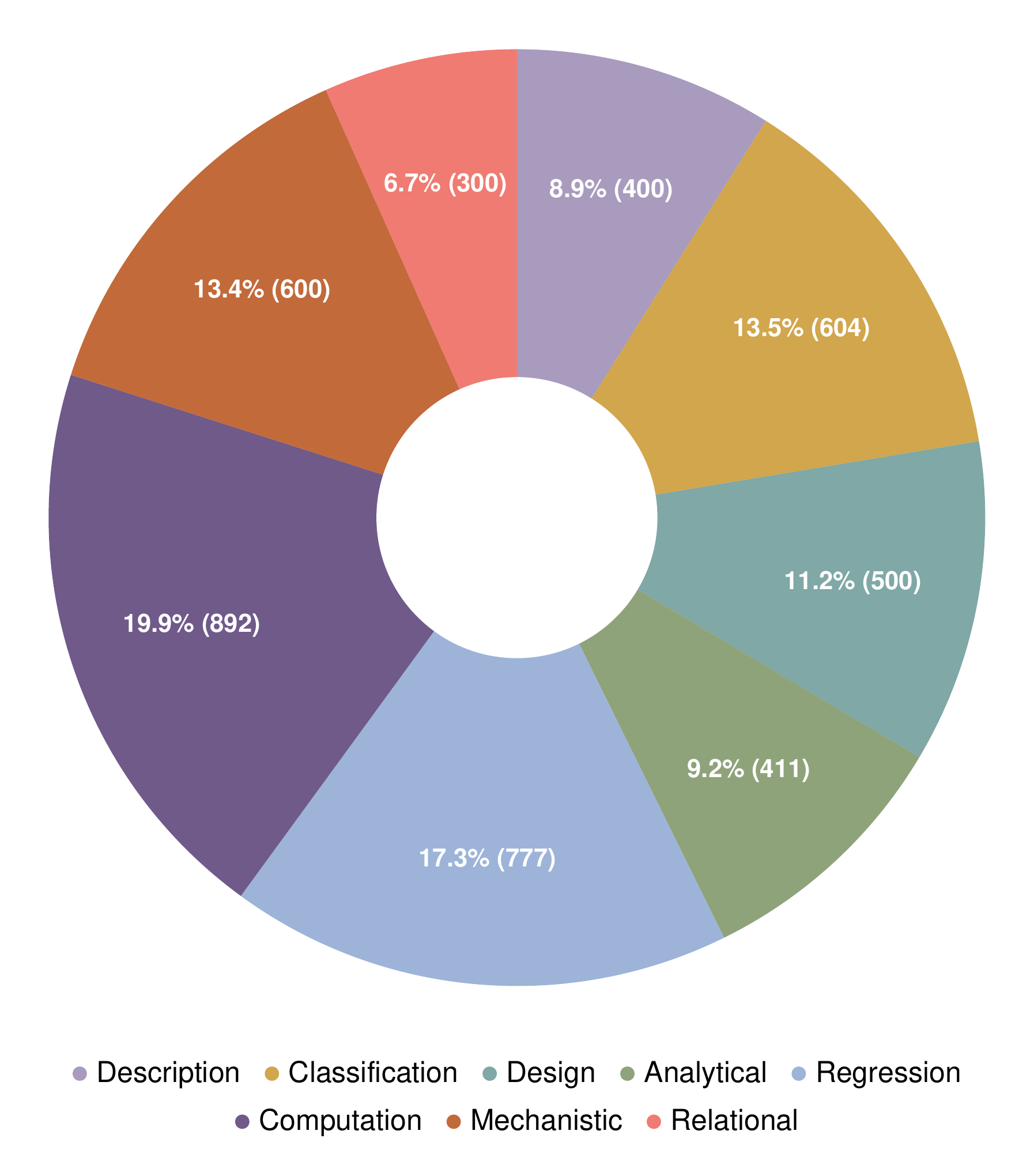}
\caption{Distribution of ChemDIRT evaluation examples across the eight chemical-reasoning categories.}
\label{fig:task_distribution}
\end{figure}

\section{The ChemDIRT Benchmark}

ChemDIRT organizes chemistry tasks according to their primary reasoning demands. Its eight categories are description, classification, design, analytical, regression, computation, mechanistic, and relational reasoning. Each is defined by the minimum chemical reasoning required for correctness rather than by prompt wording or output format. Consequently, reformulating a task may change its category when the required reasoning changes.

The taxonomy is scoped to reasoning over molecular entities, reactions, and chemical domain knowledge. Meta-level activities such as dataset curation, distributional analysis, and uncertainty estimation are excluded because their correctness does not directly depend on chemical inference.

Table \ref{tab:bench_stats} lists the 33 task settings and their evaluation counts. Representation-specific variants are shown separately if they are evaluated independently.

\begin{table*}[t]
    \centering
    \caption{ChemDIRT task--representation settings and evaluation counts ($N=4{,}484$). ``Ours'' denotes examples constructed for ChemDIRT.}
    \label{tab:bench_stats}
    \scriptsize
    \setlength{\tabcolsep}{2pt}
    \resizebox{0.99\textwidth}{!}{
    \begin{tabular}{@{}llcc@{\hspace{4pt}{\color{black!20}\vrule width 0.3pt}\hspace{4pt}}llcc@{\hspace{4pt}{\color{black!20}\vrule width 0.3pt}\hspace{4pt}}llcc@{}}
        \toprule
        Category & Subtask & Source/Test & $n$
        & Category & Subtask & Source/Test & $n$
        & Category & Subtask & Source/Test & $n$ \\
        \midrule
        Description & Carbon count & Ours & 100
        & Design & IUPAC $\rightarrow$ SMILES & Ours & 100
        & Regression & ESOL & 113 & 113 \\
        & Ring count & Ours & 100
        & & SMILES $\rightarrow$ IUPAC & Ours & 100
        & & Lipophilicity & 420 & 420 \\
        & Functional-group presence & Ours & 100
        & & Image $\rightarrow$ SMILES & Ours & 100
        & & FreeSolv & 65 & 65 \\
        & Bond-type presence & Ours & 100
        & & Image $\rightarrow$ IUPAC & Ours & 100
        & & O$_2$ permeability & 179 & 179 \\
        Classification & BACE & 152 & 152
        & & Description-based design & 100 & 100
        & Computation & Physical-unit calculation & 500 & 392 \\
        & BBBP & 204 & 204
        & Mechanistic & Forward prediction (SMILES) & 100 & 100
        & & Coefficient balancing & 500 & 100 \\
        & ClinTox & 148 & 148
        & & Forward prediction (formula) & Ours & 100
        & & Yield prediction & 300 & 300 \\
        & HIV & 4,113 & 100
        & & Forward prediction (image) & Ours & 100
        & & Molecular-weight calculation & Ours & 100 \\
        Analytical & Literature-grounded QA & 211 & 211
        & & Retrosynthesis (SMILES) & 100 & 100
        & Relational & Representation equivalence & Ours & 100 \\
        & True/false & 100 & 100
        & & Retrosynthesis (formula) & Ours & 100
        & & Chemical-similarity comparison & Ours & 100 \\
        & Open response & 100 & 100
        & & Retrosynthesis (image) & Ours & 100
        & & Molecular-weight comparison & Ours & 100 \\
        \midrule
        \multicolumn{11}{r}{\textbf{Total ChemDIRT Examples}} & \textbf{4,484} \\
        \bottomrule
    \end{tabular}
    }
    \vspace{-8px}
\end{table*}

\subsection{Task Definitions}

We define each category below and identify its ChemDIRT subtasks and reported metrics.

\subsubsection{Description}

Description tasks require intramolecular structural reasoning without empirical property inference. Formally, we define description tasks as
\[
LLM(p_{m, \tau}) = a^{(d)},
\]
where the input is a prompt $p$ containing a molecule $m$ and a structural target $\tau$, and the output is a short answer $a^{(d)}$ describing a structural attribute of the molecule. The description tasks include carbon count, ring count, functional-group presence, and bond-type presence. Carbon and ring counts require integer answers, whereas functional-group and bond-type presence require binary labels. All four tasks are evaluated using accuracy.

Molecules for these tasks are sampled randomly from QM9 \cite{ruddigkeit2012enumeration}, ZINC \cite{sterling2015zinc}, ChEMBL \cite{gaulton2012chembl}, and PubChem \cite{kim2016pubchem}. Ground-truth answers are extracted using RDKit \cite{landrum2013rdkit}. For each sampled molecule, we additionally use the PubChem API to algorithmically convert SMILES into the corresponding IUPAC name, providing an alternative linguistic representation of the same molecular input.

\subsubsection{Classification}

Classification tasks correspond to empirical molecular property inference. Given a prompt $p$ that contains a molecule $m$ and a target property $\tau$, the LLM is tasked with generating a binary classification label $y^{(c)}$, such that
\[
LLM(p_{m, \tau}) = y^{(c)}.
\]
The evaluation metric is accuracy. We include molecules from BACE, BBBP, ClinTox, and HIV as standard molecular property classification tasks following MoleculeNet \cite{wu2018moleculenet}.

\subsubsection{Design}

Design tasks require generating a molecular structure conditioned on a molecular specification. We define the design task as
\[
LLM(p_t) = m,
\]
where $p_t$ specifies a target molecule and $m$ is the generated molecule, independent of whether it is expressed in the requested SMILES or IUPAC representation. We report exact match as a strict measure and MACCS, Morgan, and RDKit (RDK) fingerprint similarity as graded measures of structural agreement, following prior recommendations for molecule generation evaluation \cite{srivastava2023beyond}. For SMILES-to-IUPAC translation, we separate structural exact match, computed after resolving the name to a structure, from literal IUPAC exact match.

We include two types of design tasks. The first comprises IUPAC-to-SMILES, SMILES-to-IUPAC, image-to-SMILES, and image-to-IUPAC translation \cite{guo2023can}. The image settings test robustness to input modality as well as representation. We construct these questions by sampling molecules from QM9 \cite{ruddigkeit2012enumeration}, ZINC \cite{sterling2015zinc}, ChEMBL \cite{gaulton2012chembl}, and PubChem \cite{kim2016pubchem}. The second is description-based design, in which a detailed textual description specifies the target molecule's structural and chemical characteristics \cite{guo2023can, fang2023mol}.

\subsubsection{Analytical}

Analytical tasks require chemical knowledge synthesis and reasoning expressed through open-form language. These tasks are defined as question--answer pairs $(q, a^{(s)})$, where
\[
LLM(p_q) = a^{(s)}.
\]
These questions require causal, deductive, inductive, or abductive inference grounded in chemical knowledge. We include literature-grounded QA, true/false questions, and short-answer open response, sourced and modified from ChemLit-QA \cite{wellawatte2025chemlit}, ChemLLMBench \cite{guo2023can}, and Mol-Instructions \cite{fang2023mol}. In literature-grounded QA, the model receives a passage as context and answers a question about that passage. Literature-grounded and open-response answers are scored using exact match and ROUGE-1, ROUGE-2, and ROUGE-L. True/false questions are scored using accuracy.

\subsubsection{Regression}

Regression tasks correspond to quantitative molecular property inference, defined as
\[
LLM(p_{m, \tau}) = y^{(r)},
\]
where the prompt $p$ contains a molecule $m$ and a target property $\tau$, and the output is a numerical label $y^{(r)}$. We include ESOL and lipophilicity from MoleculeNet \cite{wu2018moleculenet}, hydration free energy from FreeSolv \cite{mobley2014freesolv}, and O$_2$ gas permeability \cite{gautama2022suppression}. Each setting is evaluated using log root mean squared error (log RMSE), mean absolute error (MAE), relative RMSE, and RMSE. Lower values indicate better performance.

\subsubsection{Computation}

Computation tasks are governed by symbolic and calculation-based reasoning. These tasks are defined as
\[
LLM(p_q) = a^{(r)},
\]
where the question $q$ requires a numerical or coefficient-vector answer $a^{(r)}$. The computation settings are physical-unit calculation, stoichiometric coefficient balancing, reaction-yield prediction, and molecular-weight calculation, with examples drawn from ChemAlgebra \cite{valenti2024chemalgebra}, ChemistryQA \cite{ChemistryQA}, and ChemBench4K \cite{zhang2024chemllm}. We use RDKit \cite{landrum2013rdkit} to calculate reference molecular weights. Coefficient balancing is scored by accuracy while the other three settings use log RMSE, MAE, relative RMSE, and RMSE, matching the reported numerical-error metrics for regression.

\begin{table}[t]
    \centering
    \begin{tabular}{lcc}
        \toprule
        \textbf{Model} & \textbf{Full Accuracy} & \textbf{Subset Accuracy} \\
        \midrule
        ChemLLM          & 0.955 & 0.950 \\
        ChemDFM           & 0.902 & 0.900 \\
        ChemR             & 0.827 & 0.830 \\
        General-Reasoner  & 0.334 & 0.340 \\
        Falcon            & 0.043 & 0.050 \\
        Galactica         & 0.541 & 0.540 \\
        Qwen2             & 0.516 & 0.520 \\
        Qwen3             & 0.944 & 0.950 \\
        Mistral           & 0.449 & 0.450 \\
        Gemma2            & 0.958 & 0.950 \\
        Gemma3            & 0.663 & 0.660 \\
        Llama2            & 0.805 & 0.800 \\
        Llama3            & 0.357 & 0.360 \\
        DeepSeek-R1       & 0.956 & 0.950 \\
    \bottomrule
    \end{tabular}
    \vspace{1pt}
    \caption{Full-test and selected-subset HIV accuracies for reference models ($n=100$).}
    \label{tab:hiv_subset}
    \vspace{-16pt}
\end{table}

\begin{table}[h]
\centering
\footnotesize
\begin{tabular}{p{0.95\linewidth}}
\toprule
\textbf{ClinTox Classification Prompt Variants} \\
\midrule
Predict the clinical trial toxicity outcome for the molecule with SMILES: \texttt{\{SMILES\}}. \\
Did the compound represented by \texttt{\{SMILES\}} fail clinical trials due to toxicity? \\
Classify the molecule \texttt{\{SMILES\}} with respect to clinical trial toxicity. \\
Given the SMILES \texttt{\{SMILES\}}, determine whether the compound exhibits clinical trial toxicity. \\
Does the structure \texttt{\{SMILES\}} indicate failure in clinical trials due to toxicity? \\
Evaluate the clinical trial toxicity of the molecule encoded as \texttt{\{SMILES\}}. \\
For the molecule \texttt{\{SMILES\}}, return whether it failed clinical trials because of toxicity. \\
From the SMILES \texttt{\{SMILES\}}, predict the binary clinical trial toxicity label. \\
Can you assess whether \texttt{\{SMILES\}} corresponds to a clinically toxic compound? \\
What is the clinical trial toxicity classification for the compound \texttt{\{SMILES\}}? \\
\bottomrule
\end{tabular}
\vspace{-2pt}
\caption{Ten verified ClinTox prompt candidates. The reported evaluation uses the first four variants from this set, with \texttt{\{SMILES\}} replaced at inference time.}
\label{tab:prompts_clintox_example}
\vspace{-16pt}
\end{table}

\begin{figure*}[t]
    \centering
    \includegraphics[width=\textwidth]{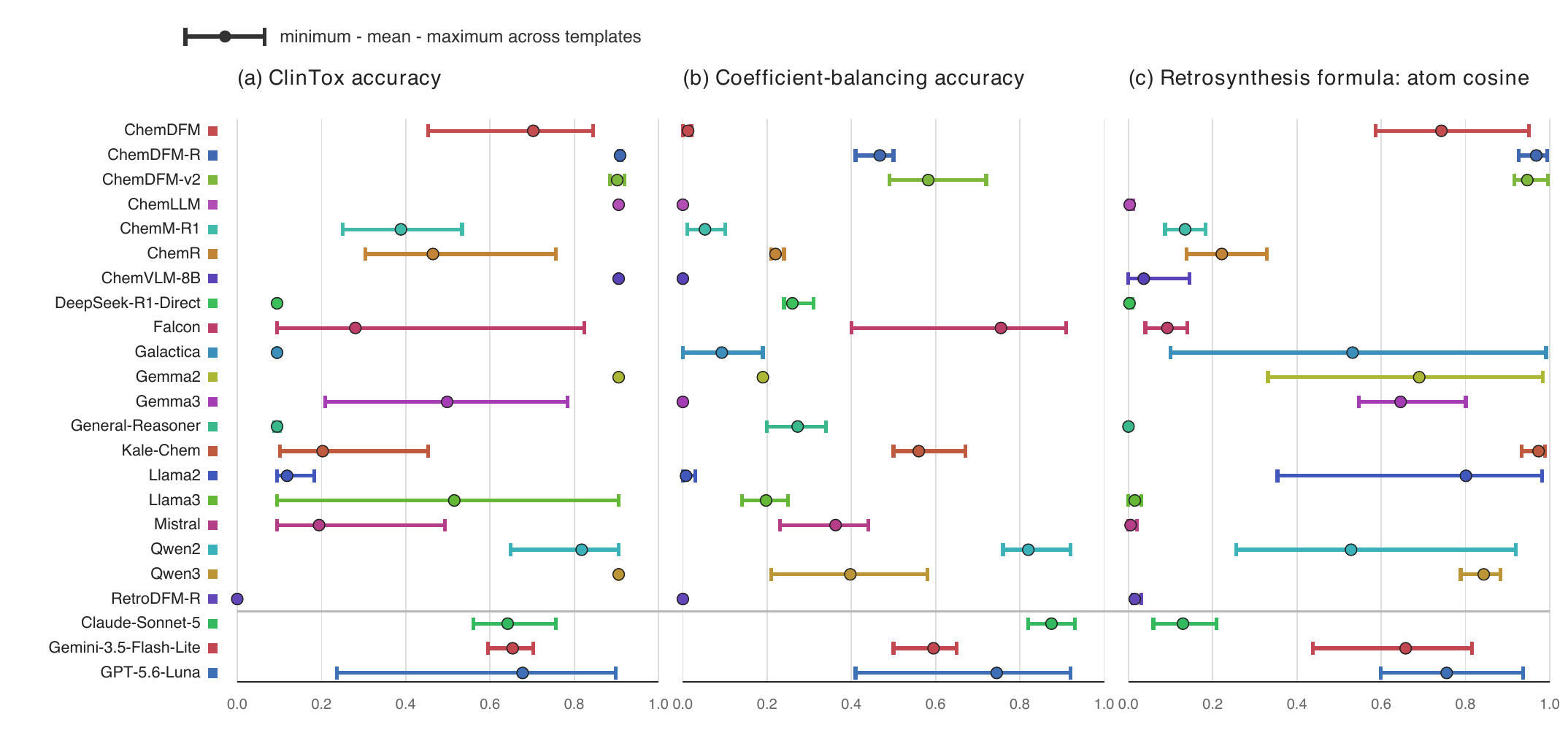}
    \vspace{-20pt}
    \caption{Variation across prompts for three tasks. Bars show the minimum and maximum. Points show the mean. Colors identify models across panels.}
    \label{fig:template_spread}
\end{figure*}

\begin{figure*}[t]
    \centering
    \includegraphics[width=\textwidth]{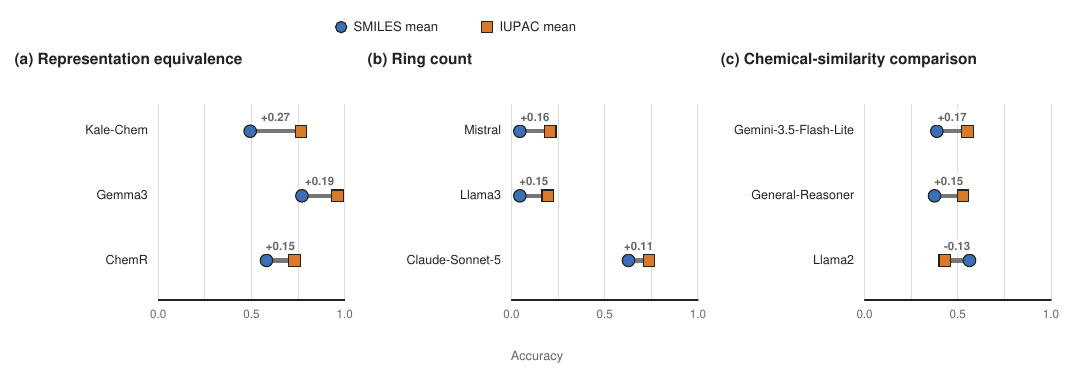}
    \vspace{-16pt}
    \caption{Representation-dependent accuracy for selected model--task pairs. Blue circles show SMILES means, orange squares show IUPAC means, and connecting bars show the labeled difference (IUPAC minus SMILES).}
    \label{fig:representation_gap}
\end{figure*}
\begin{table*}[p]
    \centering
    \caption{Prompt-varied results (mean $\pm$ SD over evaluated conditions). Panels show (a) translation, classification, and coefficient balancing, (b--c) regression, (d) mechanistic tasks, and (e) description, relational, and molecular-weight tasks.}
    \label{tab:template_task_results}
    \renewcommand{\arraystretch}{0.78}
    \begin{adjustbox}{max totalsize={0.99\textwidth}{0.84\textheight},center}
    \begin{minipage}{\textwidth}
    \centering
    \textit{(a) Translation, classification, and coefficient balancing}\par\smallskip
    \scriptsize
    \setlength{\tabcolsep}{3.5pt}
    \resizebox{0.99\textwidth}{!}{
    \begin{tabular}{|l|cccc|ccccc|c|c|c|c|c|}
        \hline
        \multirow{2}{*}{Model}
        & \multicolumn{4}{c|}{IUPAC $\rightarrow$ SMILES ($n=100$)}
        & \multicolumn{5}{c|}{SMILES $\rightarrow$ IUPAC ($n=100$)}
        & \multicolumn{1}{c|}{BACE ($n=152$)}
        & \multicolumn{1}{c|}{BBBP ($n=204$)}
        & \multicolumn{1}{c|}{ClinTox ($n=148$)}
        & \multicolumn{1}{c|}{HIV ($n=100$)}
        & \multicolumn{1}{c|}{Coefficient Balancing ($n=100$)} \\
        \cline{2-15}
        & Exact Match & MACCS & Morgan & RDK & Structural EM & IUPAC EM & MACCS & Morgan & RDK & Accuracy & Accuracy & Accuracy & Accuracy & Accuracy \\
        \hline \hline
        ChemDFM~\cite{zhao2025developing} & $0.250\pm0.012$ & $0.894\pm0.003$ & $0.651\pm0.006$ & $0.713\pm0.003$ & $0.158\pm0.045$ & $0.077\pm0.013$ & $0.822\pm0.021$ & $0.479\pm0.026$ & $0.554\pm0.025$ & $0.538\pm0.047$ & $0.567\pm0.090$ & $0.703\pm0.149$ & $0.720\pm0.028$ & $0.013\pm0.008$ \\
        ChemDFM-R~\cite{zhao2025developing} & $0.777\pm0.008$ & $0.982\pm0.002$ & $0.922\pm0.005$ & $0.949\pm0.003$ & $0.722\pm0.014$ & $0.518\pm0.004$ & $0.971\pm0.004$ & $0.897\pm0.006$ & $0.930\pm0.006$ & $0.688\pm0.029$ & $0.841\pm0.027$ & $0.909\pm0.003$ & $0.725\pm0.015$ & $0.467\pm0.037$ \\
        ChemDFM-v2~\cite{zhao2025developing} & $0.740\pm0.023$ & $0.980\pm0.002$ & $0.912\pm0.005$ & $0.940\pm0.004$ & $0.770\pm0.022$ & $0.530\pm0.000$ & $0.980\pm0.003$ & $0.914\pm0.014$ & $0.946\pm0.010$ & $0.763\pm0.022$ & $0.806\pm0.044$ & $0.902\pm0.012$ & $0.727\pm0.022$ & $0.583\pm0.089$ \\
        ChemLLM~\cite{zhang2024chemllm} & $0.022\pm0.018$ & $0.770\pm0.011$ & $0.427\pm0.017$ & $0.537\pm0.031$ & $0.000\pm0.000$ & $0.000\pm0.000$ & $0.887\pm0.000$ & $0.533\pm0.000$ & $0.813\pm0.000$ & $0.546\pm0.000$ & $0.230\pm0.000$ & $0.905\pm0.000$ & $0.700\pm0.000$ & $0.000\pm0.000$ \\
        ChemM-R1~\cite{huang2026chemm} & $0.010\pm0.007$ & $0.577\pm0.023$ & $0.246\pm0.030$ & $0.332\pm0.016$ & $0.067\pm0.094$ & $0.000\pm0.000$ & $0.531\pm0.166$ & $0.307\pm0.125$ & $0.229\pm0.230$ & $0.434\pm0.012$ & $0.688\pm0.078$ & $0.389\pm0.104$ & $0.423\pm0.094$ & $0.053\pm0.035$ \\
        ChemR~\cite{wang2026chem} & $0.728\pm0.048$ & $0.972\pm0.005$ & $0.901\pm0.011$ & $0.928\pm0.010$ & $0.721\pm0.069$ & $0.485\pm0.145$ & $0.961\pm0.014$ & $0.886\pm0.041$ & $0.906\pm0.035$ & $0.539\pm0.031$ & $0.585\pm0.118$ & $0.465\pm0.181$ & $0.480\pm0.130$ & $0.220\pm0.012$ \\
        ChemVLM-8B~\cite{li2025chemvlm} & $0.175\pm0.005$ & $0.857\pm0.007$ & $0.575\pm0.014$ & $0.668\pm0.011$ & $0.134\pm0.006$ & $0.003\pm0.004$ & $0.851\pm0.017$ & $0.567\pm0.029$ & $0.626\pm0.033$ & $0.558\pm0.016$ & $0.230\pm0.000$ & $0.905\pm0.000$ & $0.675\pm0.027$ & $0.000\pm0.000$ \\
        DeepSeek-R1-Direct~\cite{deepseekai2025r1} & \multicolumn{4}{c|}{\textit{Unable to complete}} & \multicolumn{5}{c|}{\textit{Unable to complete}} & $0.454\pm0.000$ & $0.770\pm0.000$ & $0.095\pm0.000$ & $0.300\pm0.000$ & $0.260\pm0.029$ \\
        Falcon~\cite{almazrouei2023falcon} & $0.000\pm0.000$ & $0.103\pm0.027$ & $0.024\pm0.008$ & $0.023\pm0.019$ & $0.000\pm0.000$ & $0.000\pm0.000$ & $0.117\pm0.030$ & $0.050\pm0.002$ & $0.029\pm0.009$ & $0.454\pm0.000$ & $0.643\pm0.230$ & $0.280\pm0.314$ & $0.300\pm0.000$ & $0.755\pm0.206$ \\
        Galactica~\cite{taylor2022galactica} & $0.000\pm0.000$ & $0.619\pm0.007$ & $0.240\pm0.030$ & $0.335\pm0.020$ & $0.000\pm0.000$ & $0.000\pm0.000$ & $0.218\pm0.022$ & $0.082\pm0.008$ & $0.093\pm0.015$ & $0.454\pm0.000$ & $0.770\pm0.000$ & $0.095\pm0.000$ & $0.300\pm0.000$ & $0.092\pm0.093$ \\
        Gemma2~\cite{gemmateam2024gemma2} & $0.000\pm0.000$ & $0.424\pm0.000$ & $0.165\pm0.000$ & $0.248\pm0.000$ & $0.000\pm0.000$ & $0.000\pm0.000$ & $0.253\pm0.120$ & $0.108\pm0.046$ & $0.127\pm0.085$ & $0.546\pm0.000$ & $0.230\pm0.000$ & $0.905\pm0.000$ & $0.700\pm0.000$ & $0.190\pm0.000$ \\
        Gemma3~\cite{gemmateam2025gemma3} & $0.000\pm0.000$ & $0.402\pm0.020$ & $0.135\pm0.005$ & $0.222\pm0.025$ & $0.000\pm0.000$ & $0.000\pm0.000$ & $0.375\pm0.008$ & $0.133\pm0.005$ & $0.232\pm0.010$ & $0.454\pm0.019$ & $0.232\pm0.002$ & $0.498\pm0.229$ & $0.395\pm0.097$ & $0.000\pm0.000$ \\
        General-Reasoner~\cite{ma2025general} & \multicolumn{4}{c|}{\textit{Unable to complete}} & \multicolumn{5}{c|}{\textit{Unable to complete}} & $0.454\pm0.000$ & $0.730\pm0.028$ & $0.095\pm0.005$ & $0.292\pm0.013$ & $0.273\pm0.050$ \\
        Kale-Chem~\cite{dai2025kale} & $0.117\pm0.008$ & $0.843\pm0.005$ & $0.536\pm0.006$ & $0.598\pm0.002$ & $0.148\pm0.041$ & $0.043\pm0.011$ & $0.803\pm0.016$ & $0.507\pm0.025$ & $0.561\pm0.021$ & $0.461\pm0.015$ & $0.510\pm0.258$ & $0.203\pm0.145$ & $0.338\pm0.059$ & $0.560\pm0.070$ \\
        Llama2~\cite{touvron2023llama2} & $0.000\pm0.000$ & $0.207\pm0.013$ & $0.079\pm0.015$ & $0.113\pm0.032$ & $0.000\pm0.000$ & $0.000\pm0.000$ & $0.175\pm0.018$ & $0.079\pm0.008$ & $0.069\pm0.004$ & $0.454\pm0.000$ & $0.770\pm0.000$ & $0.118\pm0.037$ & $0.300\pm0.000$ & $0.007\pm0.013$ \\
        Llama3~\cite{grattafiori2024llama3} & $0.000\pm0.000$ & $0.632\pm0.022$ & $0.215\pm0.008$ & $0.362\pm0.011$ & $0.000\pm0.000$ & $0.000\pm0.000$ & $0.481\pm0.018$ & $0.160\pm0.006$ & $0.238\pm0.050$ & $0.503\pm0.079$ & $0.540\pm0.240$ & $0.515\pm0.384$ & $0.435\pm0.156$ & $0.198\pm0.039$ \\
        Mistral~\cite{jiang2023mistral} & $0.000\pm0.000$ & $0.403\pm0.033$ & $0.120\pm0.012$ & $0.194\pm0.028$ & $0.000\pm0.000$ & $0.000\pm0.000$ & $0.338\pm9.96\!\times\!10^{-4}$ & $0.122\pm0.005$ & $0.192\pm0.007$ & $0.475\pm0.041$ & $0.752\pm0.019$ & $0.194\pm0.173$ & $0.400\pm0.173$ & $0.362\pm0.079$ \\
        Qwen2~\cite{yang2024qwen2} & $0.000\pm0.000$ & $0.368\pm0.020$ & $0.124\pm0.008$ & $0.232\pm0.012$ & $0.000\pm0.000$ & $0.000\pm0.000$ & $0.278\pm0.007$ & $0.096\pm0.004$ & $0.153\pm0.013$ & $0.551\pm0.009$ & $0.230\pm0.000$ & $0.818\pm0.103$ & $0.642\pm0.088$ & $0.820\pm0.060$ \\
        Qwen3~\cite{qwenteam2025qwen3} & $0.000\pm0.000$ & $0.517\pm0.011$ & $0.170\pm0.008$ & $0.283\pm0.006$ & $0.000\pm0.000$ & $0.000\pm0.000$ & $0.440\pm0.012$ & $0.153\pm0.004$ & $0.218\pm0.012$ & $0.535\pm0.020$ & $0.232\pm0.002$ & $0.905\pm0.000$ & $0.702\pm0.004$ & $0.397\pm0.132$ \\
        RetroDFM-R~\cite{zhang2025retrodfmr} & $0.018\pm0.019$ & $0.710\pm0.021$ & $0.426\pm0.014$ & $0.513\pm0.011$ & \multicolumn{5}{c|}{\textit{Unable to complete}} & $0.457\pm0.006$ & $0.772\pm0.004$ & $0.150\pm0.013$ & $0.305\pm0.005$ & $0.000\pm0.000$ \\
        \hline
        Claude-Sonnet-5~\cite{anthropic2026models} & $0.487\pm0.033$ & $0.901\pm0.040$ & $0.812\pm0.033$ & $0.834\pm0.037$ & $0.548\pm0.032$ & $0.180\pm0.016$ & $0.960\pm0.008$ & $0.836\pm0.013$ & $0.874\pm0.010$ & $0.523\pm0.028$ & $0.799\pm0.037$ & $0.642\pm0.072$ & $0.725\pm0.023$ & $0.875\pm0.043$ \\
        Gemini-3.5-Flash-Lite~\cite{google2026geminimodels} & $0.145\pm0.017$ & $0.881\pm0.013$ & $0.618\pm0.019$ & $0.675\pm0.024$ & $0.172\pm0.035$ & $0.038\pm0.004$ & $0.873\pm0.021$ & $0.585\pm0.031$ & $0.655\pm0.030$ & $0.484\pm0.012$ & $0.809\pm0.023$ & $0.654\pm0.040$ & $0.593\pm0.057$ & $0.595\pm0.057$ \\
        GPT-5.6-Luna~\cite{openai2026models} & $0.158\pm0.011$ & $0.885\pm0.012$ & $0.614\pm0.013$ & $0.678\pm0.008$ & $0.090\pm0.020$ & $0.025\pm0.005$ & $0.833\pm0.002$ & $0.478\pm0.016$ & $0.567\pm0.009$ & $0.543\pm0.078$ & $0.525\pm0.089$ & $0.677\pm0.265$ & $0.613\pm0.118$ & $0.745\pm0.203$ \\
        \hline
    \end{tabular}
    }
    \par\vspace{1pt}
    \textit{(b) Regression: ESOL and lipophilicity}\par\smallskip
    \scriptsize
    \setlength{\tabcolsep}{3.5pt}
    \resizebox{0.99\textwidth}{!}{
    \begin{tabular}{|l|cccc|cccc|}
        \hline
        \multirow{2}{*}{Model}
        & \multicolumn{4}{c|}{ESOL ($n=113$)}
        & \multicolumn{4}{c|}{Lipophilicity ($n=420$)} \\
        \cline{2-9}
        & Log RMSE & MAE & Relative RMSE & RMSE & Log RMSE & MAE & Relative RMSE & RMSE \\
        \hline \hline
        ChemDFM & $3.269\pm0.145$ & $467.239\pm108.379$ & $2.54\!\times\!10^{3}\pm1.76\!\times\!10^{3}$ & $870.280\pm186.688$ & $1.919\pm0.236$ & $300.031\pm102.987$ & $955.242\pm479.278$ & $558.706\pm147.705$ \\
        ChemDFM-R & $0.467\pm0.142$ & $3.602\pm0.196$ & $2.064\pm1.155$ & $4.288\pm0.198$ & $0.891\pm0.147$ & $2.115\pm0.155$ & $1.747\pm0.189$ & $2.257\pm0.146$ \\
        ChemDFM-v2 & $0.761\pm0.367$ & $3.733\pm0.143$ & $3.869\pm1.600$ & $4.401\pm0.157$ & $1.017\pm0.130$ & $2.030\pm0.053$ & $1.336\pm0.146$ & $2.206\pm0.055$ \\
        ChemLLM & $0.986\pm0.260$ & $3.623\pm0.053$ & $1.712\pm1.211$ & $4.328\pm0.118$ & $0.513\pm0.107$ & $5.053\pm0.686$ & $6.601\pm1.931$ & $5.840\pm0.306$ \\
        ChemM-R1 & $17.364\pm23.085$ & $9.32\!\times\!10^{125}\pm1.16\!\times\!10^{126}$ & $3.59\!\times\!10^{126}\pm5.76\!\times\!10^{126}$ & $3.30\!\times\!10^{126}\pm3.31\!\times\!10^{126}$ & $13.824\pm7.875$ & $2.14\!\times\!10^{125}\pm2.40\!\times\!10^{125}$ & $5.95\!\times\!10^{125}\pm6.18\!\times\!10^{125}$ & $1.39\!\times\!10^{126}\pm1.57\!\times\!10^{126}$ \\
        ChemR & $1.306\pm0.515$ & $2.16\!\times\!10^{25}\pm3.74\!\times\!10^{25}$ & $3.37\!\times\!10^{25}\pm5.84\!\times\!10^{25}$ & $2.29\!\times\!10^{26}\pm3.97\!\times\!10^{26}$ & $1.446\pm0.476$ & $1.20\!\times\!10^{34}\pm2.07\!\times\!10^{34}$ & $5.84\!\times\!10^{34}\pm1.01\!\times\!10^{35}$ & $2.45\!\times\!10^{35}\pm4.25\!\times\!10^{35}$ \\
        ChemVLM-8B & $0.816\pm0.403$ & $3.598\pm0.047$ & $1.093\pm0.113$ & $4.266\pm0.041$ & $0.294\pm0.021$ & $0.919\pm0.064$ & $5.467\pm0.666$ & $1.133\pm0.052$ \\
        DeepSeek-R1-Direct & $1.596\pm0.848$ & $4.514\pm0.110$ & $5.951\pm0.737$ & $5.109\pm0.080$ & $0.332\pm0.038$ & $1.222\pm0.144$ & $4.406\pm0.479$ & $1.422\pm0.144$ \\
        Falcon & $0.771\pm0.616$ & $4.034\pm0.387$ & $2.392\pm1.731$ & $4.691\pm0.402$ & $0.432\pm0.007$ & $1.992\pm0.276$ & $1.649\pm0.759$ & $2.285\pm0.260$ \\
        Galactica & $2.283\pm1.682$ & $3.840\pm0.408$ & $2.212\pm1.807$ & $4.484\pm0.361$ & $0.408\pm0.019$ & $1.594\pm0.267$ & $3.363\pm0.206$ & $1.833\pm0.330$ \\
        Gemma2 & $3.369\pm1.537$ & $4.117\pm0.479$ & $2.738\pm1.814$ & $4.814\pm0.499$ & $0.394\pm0.022$ & $1.437\pm0.063$ & $3.163\pm0.240$ & $1.634\pm0.064$ \\
        Gemma3 & $0.995\pm0.531$ & $6.139\pm3.955$ & $15.105\pm20.433$ & $8.963\pm7.788$ & $0.453\pm0.117$ & $1.984\pm0.794$ & $6.883\pm2.724$ & $2.685\pm1.165$ \\
        General-Reasoner & $1.625\pm0.445$ & $11.557\pm6.373$ & $53.114\pm46.797$ & $35.612\pm38.406$ & $0.439\pm0.018$ & $2.605\pm0.424$ & $7.733\pm2.207$ & $5.940\pm0.819$ \\
        Kale-Chem & $1.420\pm0.367$ & $3.631\pm0.062$ & $1.184\pm0.049$ & $4.283\pm0.057$ & $0.280\pm0.009$ & $0.981\pm0.073$ & $5.725\pm0.826$ & $1.183\pm0.062$ \\
        Llama2 & $1.959\pm1.335$ & $4.211\pm0.383$ & $3.470\pm1.890$ & $4.812\pm0.344$ & $0.406\pm0.047$ & $1.495\pm0.077$ & $3.310\pm0.728$ & $1.800\pm0.103$ \\
        Llama3 & $1.853\pm0.767$ & $10.165\pm6.024$ & $46.041\pm42.090$ & $38.598\pm34.490$ & $0.479\pm0.125$ & $2.114\pm0.351$ & $3.077\pm1.272$ & $2.393\pm0.384$ \\
        Mistral & $4.703\pm1.497$ & $69.986\pm73.441$ & $166.570\pm164.718$ & $374.208\pm574.957$ & $0.348\pm0.029$ & $1.303\pm0.104$ & $4.540\pm0.570$ & $1.608\pm0.242$ \\
        Qwen2 & $1.583\pm1.220$ & $3.468\pm0.333$ & $7.568\pm10.098$ & $4.109\pm0.344$ & $0.588\pm0.162$ & $3.756\pm0.629$ & $8.289\pm1.629$ & $3.965\pm0.595$ \\
        Qwen3 & $1.905\pm0.297$ & $5.415\pm3.560$ & $25.830\pm40.873$ & $10.142\pm10.555$ & $0.496\pm0.203$ & $1.992\pm0.695$ & $5.097\pm2.589$ & $2.310\pm0.621$ \\
        RetroDFM-R & \multicolumn{4}{c|}{\textit{Unable to complete}} & $1.188\pm0.205$ & $1.57\!\times\!10^{12}\pm2.73\!\times\!10^{12}$ & $1.83\!\times\!10^{13}\pm3.17\!\times\!10^{13}$ & $3.17\!\times\!10^{13}\pm5.49\!\times\!10^{13}$ \\
        \hline
        Claude-Sonnet-5 & $0.396\pm0.096$ & $0.987\pm0.564$ & $65.158\pm109.087$ & $2.036\pm2.034$ & $0.278\pm0.009$ & $0.927\pm0.097$ & $5.794\pm0.146$ & $1.259\pm0.155$ \\
        Gemini-3.5-Flash-Lite & $1.892\pm1.352$ & $539.823\pm850.388$ & $1.44\!\times\!10^{4}\pm2.44\!\times\!10^{4}$ & $4.43\!\times\!10^{3}\pm7.34\!\times\!10^{3}$ & $0.322\pm0.010$ & $1.176\pm0.090$ & $9.328\pm1.077$ & $1.592\pm0.084$ \\
        GPT-5.6-Luna & $1.051\pm0.710$ & $29.384\pm46.244$ & $838.976\pm1.45\!\times\!10^{3}$ & $245.542\pm419.288$ & $0.299\pm0.003$ & $1.006\pm0.021$ & $7.470\pm0.914$ & $1.328\pm0.020$ \\
        \hline
    \end{tabular}
    }
    \par\vspace{1pt}
    \textit{(c) Regression: FreeSolv and O$_2$ permeability}\par\smallskip
    \scriptsize
    \setlength{\tabcolsep}{3.5pt}
    \resizebox{0.99\textwidth}{!}{
    \begin{tabular}{|l|cccc|cccc|}
        \hline
        \multirow{2}{*}{Model}
        & \multicolumn{4}{c|}{FreeSolv ($n=65$)}
        & \multicolumn{4}{c|}{O$_2$ Permeability ($n=179$)} \\
        \cline{2-9}
        & Log RMSE & MAE & Relative RMSE & RMSE & Log RMSE & MAE & Relative RMSE & RMSE \\
        \hline \hline
        ChemDFM & $3.015\pm0.239$ & $431.903\pm163.933$ & $6.89\!\times\!10^{4}\pm2.90\!\times\!10^{4}$ & $461.251\pm154.836$ & $3.395\pm0.646$ & $6.50\!\times\!10^{15}\pm1.13\!\times\!10^{16}$ & $1.29\!\times\!10^{16}\pm2.23\!\times\!10^{16}$ & $6.70\!\times\!10^{16}\pm1.16\!\times\!10^{17}$ \\
        ChemDFM-R & $0.768\pm0.271$ & $0.530\pm0.217$ & $17.258\pm15.379$ & $0.790\pm0.287$ & $3.855\pm0.135$ & $358.945\pm0.003$ & $1.055\pm0.068$ & $1.82\!\times\!10^{3}\pm3.00\!\times\!10^{-4}$ \\
        ChemDFM-v2 & $1.123\pm0.285$ & $0.738\pm0.152$ & $17.683\pm14.237$ & $1.017\pm0.230$ & $3.926\pm0.047$ & $358.913\pm0.038$ & $0.996\pm0.003$ & $1.82\!\times\!10^{3}\pm0.002$ \\
        ChemLLM & $0.947\pm0.000$ & $11.331\pm7.437$ & $967.325\pm1.34\!\times\!10^{3}$ & $19.390\pm10.843$ & $3.101\pm0.561$ & $372.004\pm11.597$ & $207.927\pm221.118$ & $1.83\!\times\!10^{3}\pm9.503$ \\
        ChemM-R1 & $3.806\pm0.867$ & $7.03\!\times\!10^{15}\pm1.22\!\times\!10^{16}$ & $8.40\!\times\!10^{16}\pm1.45\!\times\!10^{17}$ & $5.45\!\times\!10^{16}\pm9.43\!\times\!10^{16}$ & $17.538\pm5.472$ & $1.89\!\times\!10^{182}\pm3.27\!\times\!10^{182}$ & $7.02\!\times\!10^{182}\pm1.22\!\times\!10^{183}$ & $2.51\!\times\!10^{183}\pm4.35\!\times\!10^{183}$ \\
        ChemR & $2.346\pm0.200$ & $1.03\!\times\!10^{3}\pm1.50\!\times\!10^{3}$ & $6.04\!\times\!10^{3}\pm5.16\!\times\!10^{3}$ & $5.34\!\times\!10^{3}\pm8.57\!\times\!10^{3}$ & $5.623\pm1.615$ & $4.62\!\times\!10^{21}\pm7.66\!\times\!10^{21}$ & $1.31\!\times\!10^{21}\pm2.25\!\times\!10^{21}$ & $6.19\!\times\!10^{22}\pm1.02\!\times\!10^{23}$ \\
        ChemVLM-8B & $1.450\pm0.000$ & $7.195\pm8.879$ & $7.97\!\times\!10^{3}\pm1.37\!\times\!10^{4}$ & $24.188\pm26.420$ & $1.986\pm1.102$ & $358.934\pm0.007$ & $0.995\pm0.006$ & $1.82\!\times\!10^{3}\pm0.001$ \\
        DeepSeek-R1-Direct & $4.530\pm4.464$ & $1.226\pm0.193$ & $164.381\pm97.818$ & $1.716\pm0.252$ & $6.124\pm3.644$ & $3.38\!\times\!10^{16}\pm3.21\!\times\!10^{16}$ & $2.23\!\times\!10^{17}\pm2.25\!\times\!10^{17}$ & $1.58\!\times\!10^{17}\pm1.10\!\times\!10^{17}$ \\
        Falcon & $1.179\pm1.138$ & $1.943\pm1.339$ & $119.898\pm115.155$ & $3.176\pm2.347$ & $1.439\pm0.101$ & $358.604\pm0.284$ & $8.387\pm6.848$ & $1.82\!\times\!10^{3}\pm0.096$ \\
        Galactica & $0.754\pm0.000$ & $5.961\pm4.106$ & $902.832\pm720.299$ & $6.546\pm3.682$ & $9.721\pm11.783$ & $1.40\!\times\!10^{125}\pm2.42\!\times\!10^{125}$ & $2.39\!\times\!10^{126}\pm4.15\!\times\!10^{126}$ & $5.91\!\times\!10^{125}\pm1.02\!\times\!10^{126}$ \\
        Gemma2 & $16.313\pm23.899$ & $5.77\!\times\!10^{125}\pm9.99\!\times\!10^{125}$ & $2.50\!\times\!10^{126}\pm4.33\!\times\!10^{126}$ & $1.20\!\times\!10^{126}\pm2.08\!\times\!10^{126}$ & $1.754\pm0.765$ & $358.256\pm0.053$ & $16.447\pm1.477$ & $1.82\!\times\!10^{3}\pm0.022$ \\
        Gemma3 & $5.745\pm5.071$ & $42.888\pm38.121$ & $1.30\!\times\!10^{4}\pm1.13\!\times\!10^{4}$ & $59.939\pm57.829$ & $44.744\pm40.472$ & $2.69\!\times\!10^{126}\pm4.14\!\times\!10^{126}$ & $5.31\!\times\!10^{127}\pm5.11\!\times\!10^{127}$ & $4.18\!\times\!10^{126}\pm4.20\!\times\!10^{126}$ \\
        General-Reasoner & $1.253\pm0.388$ & $12.839\pm10.089$ & $663.265\pm662.819$ & $29.800\pm20.731$ & $1.867\pm0.231$ & $486.039\pm123.261$ & $850.657\pm239.061$ & $1.83\!\times\!10^{3}\pm679.184$ \\
        Kale-Chem & $0.510\pm0.208$ & $9.254\pm9.575$ & $46.147\pm29.036$ & $27.052\pm25.152$ & $4.893\pm0.132$ & $358.941\pm5.15\!\times\!10^{-5}$ & $1.000\pm1.27\!\times\!10^{-4}$ & $1.82\!\times\!10^{3}\pm1.05\!\times\!10^{-5}$ \\
        Llama2 & $0.730\pm0.000$ & $0.987\pm0.086$ & $40.746\pm68.841$ & $1.391\pm0.143$ & $27.016\pm32.171$ & $1.10\!\times\!10^{126}\pm1.69\!\times\!10^{126}$ & $3.49\!\times\!10^{126}\pm4.72\!\times\!10^{126}$ & $2.08\!\times\!10^{126}\pm2.59\!\times\!10^{126}$ \\
        Llama3 & $0.655\pm0.497$ & $4.760\pm6.006$ & $108.001\pm46.874$ & $12.863\pm19.530$ & $2.552\pm1.180$ & $4.88\!\times\!10^{5}\pm8.45\!\times\!10^{5}$ & $1.37\!\times\!10^{6}\pm2.36\!\times\!10^{6}$ & $1.40\!\times\!10^{6}\pm2.42\!\times\!10^{6}$ \\
        Mistral & $2.520\pm1.612$ & $31.740\pm27.626$ & $3.38\!\times\!10^{3}\pm1.90\!\times\!10^{3}$ & $41.388\pm42.527$ & $71.319\pm30.945$ & $4.89\!\times\!10^{126}\pm4.82\!\times\!10^{126}$ & $6.48\!\times\!10^{127}\pm6.07\!\times\!10^{127}$ & $7.44\!\times\!10^{126}\pm5.22\!\times\!10^{126}$ \\
        Qwen2 & \multicolumn{4}{c|}{\textit{Unable to complete}} & $4.284\pm1.861$ & $3.14\!\times\!10^{7}\pm5.01\!\times\!10^{7}$ & $1.49\!\times\!10^{8}\pm2.52\!\times\!10^{8}$ & $1.18\!\times\!10^{8}\pm1.56\!\times\!10^{8}$ \\
        Qwen3 & $2.175\pm0.000$ & $15.707\pm15.351$ & $2.21\!\times\!10^{3}\pm2.77\!\times\!10^{3}$ & $28.860\pm25.399$ & $3.847\pm1.003$ & $359.812\pm0.952$ & $54.016\pm76.537$ & $1.82\!\times\!10^{3}\pm0.766$ \\
        RetroDFM-R & $0.198\pm0.000$ & $0.311\pm0.000$ & $0.366\pm0.000$ & $0.311\pm0.000$ & $0.461\pm0.000$ & $1.890\pm0.000$ & $0.654\pm0.000$ & $1.890\pm0.000$ \\
        \hline
        Claude-Sonnet-5 & $0.475\pm0.436$ & $4.829\pm0.992$ & $599.838\pm20.971$ & $7.437\pm3.434$ & $1.145\pm0.039$ & $366.368\pm0.513$ & $104.889\pm115.303$ & $1.82\!\times\!10^{3}\pm0.378$ \\
        Gemini-3.5-Flash-Lite & $0.670\pm0.519$ & $10.657\pm4.348$ & $1.48\!\times\!10^{3}\pm1.38\!\times\!10^{3}$ & $14.699\pm5.116$ & $1.162\pm0.082$ & $503.858\pm84.337$ & $253.386\pm246.040$ & $2.05\!\times\!10^{3}\pm107.004$ \\
        GPT-5.6-Luna & $0.291\pm0.071$ & $8.395\pm4.182$ & $638.138\pm61.458$ & $19.699\pm13.976$ & $2.480\pm0.938$ & $524.793\pm273.093$ & $2.27\!\times\!10^{4}\pm3.91\!\times\!10^{4}$ & $3.29\!\times\!10^{3}\pm2.56\!\times\!10^{3}$ \\
        \hline
    \end{tabular}
    }
    \par\vspace{1pt}
    \textit{(d) Mechanistic tasks}\par\smallskip
    \scriptsize
    \setlength{\tabcolsep}{3.5pt}
    \resizebox{0.99\textwidth}{!}{
    \begin{tabular}{|l|cccc|ccc|cccc|ccc|}
        \hline
        \multirow{2}{*}{Model}
        & \multicolumn{4}{c|}{Forward (SMILES) ($n=100$)}
        & \multicolumn{3}{c|}{Forward (Formula) ($n=100$)}
        & \multicolumn{4}{c|}{Retrosynthesis (SMILES) ($n=100$)}
        & \multicolumn{3}{c|}{Retrosynthesis (Formula) ($n=100$)} \\
        \cline{2-15}
        & Exact Match & MACCS & Morgan & RDK & Exact Match & Atom Cosine & Atom F1 & Exact Match & MACCS & Morgan & RDK & Exact Match & Atom Cosine & Atom F1 \\
        \hline \hline
        ChemDFM & $0.470\pm0.049$ & $0.846\pm0.015$ & $0.740\pm0.020$ & $0.782\pm0.014$ & $0.000\pm0.000$ & $0.917\pm0.094$ & $0.769\pm0.074$ & $0.122\pm0.025$ & $0.793\pm0.005$ & $0.629\pm0.016$ & $0.742\pm0.010$ & $0.000\pm0.000$ & $0.743\pm0.142$ & $0.657\pm0.098$ \\
        ChemDFM-R & $0.680\pm0.010$ & $0.911\pm9.36\!\times\!10^{-4}$ & $0.834\pm0.003$ & $0.861\pm0.004$ & $0.000\pm0.000$ & $0.995\pm1.38\!\times\!10^{-4}$ & $0.888\pm0.005$ & $0.497\pm0.015$ & $0.882\pm0.005$ & $0.780\pm0.007$ & $0.838\pm0.005$ & $0.000\pm0.000$ & $0.968\pm0.026$ & $0.908\pm0.025$ \\
        ChemDFM-v2 & $0.620\pm0.033$ & $0.888\pm0.012$ & $0.807\pm0.011$ & $0.837\pm0.017$ & $0.000\pm0.000$ & $0.995\pm1.76\!\times\!10^{-4}$ & $0.901\pm0.006$ & $0.425\pm0.011$ & $0.860\pm0.002$ & $0.738\pm0.002$ & $0.796\pm0.003$ & $0.000\pm0.000$ & $0.947\pm0.033$ & $0.885\pm0.039$ \\
        ChemLLM & $0.000\pm0.000$ & $0.700\pm0.000$ & $0.521\pm0.000$ & $0.709\pm0.000$ & $0.000\pm0.000$ & $0.007\pm0.013$ & $0.005\pm0.008$ & $0.000\pm0.000$ & $0.734\pm0.056$ & $0.535\pm0.081$ & $0.620\pm0.100$ & $0.000\pm0.000$ & $0.003\pm0.005$ & $0.003\pm0.004$ \\
        ChemM-R1 & $0.000\pm0.000$ & $0.604\pm0.033$ & $0.357\pm0.050$ & $0.398\pm0.048$ & $0.000\pm0.000$ & $0.153\pm0.016$ & $0.118\pm0.020$ & $0.000\pm0.000$ & $0.690\pm0.038$ & $0.475\pm0.047$ & $0.563\pm0.051$ & $0.000\pm0.000$ & $0.134\pm0.047$ & $0.107\pm0.043$ \\
        ChemR & $0.645\pm0.026$ & $0.910\pm0.009$ & $0.848\pm0.010$ & $0.888\pm0.004$ & $0.003\pm0.004$ & $0.361\pm0.085$ & $0.244\pm0.068$ & $0.208\pm0.120$ & $0.816\pm0.032$ & $0.658\pm0.050$ & $0.746\pm0.027$ & $0.000\pm0.000$ & $0.222\pm0.071$ & $0.168\pm0.056$ \\
        ChemVLM-8B & $0.003\pm0.004$ & $0.829\pm0.023$ & $0.528\pm0.043$ & $0.689\pm0.003$ & $0.000\pm0.000$ & $0.015\pm0.026$ & $0.014\pm0.024$ & $0.035\pm0.009$ & $0.600\pm0.275$ & $0.484\pm0.222$ & $0.563\pm0.262$ & $0.000\pm0.000$ & $0.036\pm0.063$ & $0.031\pm0.053$ \\
        DeepSeek-R1-Direct & $0.000\pm0.000$ & $0.544\pm0.222$ & $0.378\pm0.003$ & $0.340\pm0.128$ & $0.000\pm0.000$ & $0.000\pm0.000$ & $0.000\pm0.000$ & $0.000\pm0.000$ & $0.211\pm0.060$ & $0.159\pm0.063$ & $0.137\pm0.117$ & $0.000\pm0.000$ & $0.003\pm0.004$ & $0.003\pm0.004$ \\
        Falcon & $0.000\pm0.000$ & $0.441\pm0.000$ & $0.318\pm0.000$ & $0.355\pm0.000$ & $0.000\pm0.000$ & $0.032\pm0.055$ & $0.021\pm0.036$ & $0.000\pm0.000$ & $0.470\pm0.202$ & $0.357\pm0.188$ & $0.500\pm0.307$ & $0.000\pm0.000$ & $0.092\pm0.048$ & $0.088\pm0.050$ \\
        Galactica & $0.000\pm0.000$ & $0.579\pm0.262$ & $0.485\pm0.260$ & $0.574\pm0.277$ & $0.000\pm0.000$ & $0.423\pm0.390$ & $0.335\pm0.303$ & $0.000\pm0.000$ & $0.765\pm0.083$ & $0.613\pm0.091$ & $0.736\pm0.111$ & $0.000\pm0.000$ & $0.532\pm0.332$ & $0.472\pm0.269$ \\
        Gemma2 & $0.000\pm0.000$ & $0.686\pm0.019$ & $0.515\pm0.025$ & $0.648\pm0.050$ & $0.000\pm0.000$ & $0.598\pm0.227$ & $0.534\pm0.207$ & $0.000\pm0.000$ & $0.649\pm0.125$ & $0.468\pm0.109$ & $0.532\pm0.111$ & $0.000\pm0.000$ & $0.690\pm0.262$ & $0.608\pm0.240$ \\
        Gemma3 & $0.000\pm0.000$ & $0.547\pm0.023$ & $0.355\pm0.019$ & $0.450\pm0.027$ & $0.000\pm0.000$ & $0.721\pm0.029$ & $0.611\pm0.016$ & $0.000\pm0.000$ & $0.640\pm0.057$ & $0.445\pm0.063$ & $0.546\pm0.068$ & $0.000\pm0.000$ & $0.646\pm0.104$ & $0.577\pm0.087$ \\
        General-Reasoner & $0.000\pm0.000$ & $0.647\pm0.000$ & $0.373\pm0.000$ & $0.445\pm0.000$ & $0.000\pm0.000$ & $0.000\pm0.000$ & $0.000\pm0.000$ & \multicolumn{4}{c|}{\textit{Unable to complete}} & $0.000\pm0.000$ & $0.000\pm0.000$ & $0.000\pm0.000$ \\
        Kale-Chem & $0.185\pm0.011$ & $0.771\pm0.005$ & $0.600\pm0.009$ & $0.663\pm0.010$ & $0.000\pm0.000$ & $0.983\pm0.011$ & $0.699\pm0.043$ & $0.020\pm0.007$ & $0.678\pm0.045$ & $0.476\pm0.033$ & $0.571\pm0.047$ & $0.000\pm0.000$ & $0.973\pm0.023$ & $0.759\pm0.030$ \\
        Llama2 & $0.000\pm0.000$ & $0.516\pm0.048$ & $0.373\pm0.046$ & $0.458\pm0.057$ & $0.000\pm0.000$ & $0.677\pm0.306$ & $0.551\pm0.248$ & $0.000\pm0.000$ & $0.657\pm0.104$ & $0.502\pm0.091$ & $0.583\pm0.110$ & $0.000\pm0.000$ & $0.801\pm0.261$ & $0.739\pm0.234$ \\
        Llama3 & $0.000\pm0.000$ & $0.093\pm0.091$ & $0.062\pm0.070$ & $0.078\pm0.091$ & $0.000\pm0.000$ & $0.000\pm0.000$ & $0.000\pm0.000$ & $0.000\pm0.000$ & $0.595\pm0.119$ & $0.363\pm0.143$ & $0.462\pm0.148$ & $0.000\pm0.000$ & $0.015\pm0.015$ & $0.015\pm0.015$ \\
        Mistral & $0.000\pm0.000$ & $0.716\pm0.000$ & $0.546\pm0.000$ & $0.633\pm0.000$ & $0.000\pm0.000$ & $0.002\pm0.004$ & $0.002\pm0.004$ & $0.000\pm0.000$ & $0.131\pm0.118$ & $0.041\pm0.041$ & $0.055\pm0.055$ & $0.000\pm0.000$ & $0.005\pm0.009$ & $0.005\pm0.008$ \\
        Qwen2 & $0.000\pm0.000$ & $0.588\pm0.069$ & $0.397\pm0.074$ & $0.498\pm0.077$ & $0.000\pm0.000$ & $0.685\pm0.138$ & $0.636\pm0.140$ & $0.000\pm0.000$ & $0.557\pm0.141$ & $0.372\pm0.145$ & $0.453\pm0.164$ & $0.000\pm0.000$ & $0.528\pm0.260$ & $0.495\pm0.265$ \\
        Qwen3 & $0.005\pm0.005$ & $0.617\pm0.031$ & $0.444\pm0.026$ & $0.537\pm0.038$ & $0.000\pm0.000$ & $0.629\pm0.112$ & $0.538\pm0.093$ & $0.000\pm0.000$ & $0.696\pm0.035$ & $0.486\pm0.036$ & $0.589\pm0.045$ & $0.000\pm0.000$ & $0.843\pm0.035$ & $0.782\pm0.017$ \\
        RetroDFM-R & $0.000\pm0.000$ & $0.692\pm0.006$ & $0.500\pm0.009$ & $0.634\pm0.009$ & $0.000\pm0.000$ & $0.020\pm0.024$ & $0.020\pm0.024$ & $0.685\pm0.023$ & $0.935\pm0.010$ & $0.874\pm0.017$ & $0.912\pm0.012$ & $0.000\pm0.000$ & $0.015\pm0.009$ & $0.014\pm0.009$ \\
        \hline
        Claude-Sonnet-5 & $0.497\pm0.044$ & $0.768\pm0.029$ & $0.689\pm0.041$ & $0.723\pm0.034$ & $0.000\pm0.000$ & $0.259\pm0.146$ & $0.192\pm0.118$ & $0.113\pm0.025$ & $0.349\pm0.122$ & $0.280\pm0.093$ & $0.296\pm0.107$ & $0.000\pm0.000$ & $0.129\pm0.070$ & $0.118\pm0.070$ \\
        Gemini-3.5-Flash-Lite & $0.083\pm0.024$ & $0.701\pm0.010$ & $0.500\pm0.014$ & $0.589\pm0.009$ & $0.000\pm0.000$ & $0.554\pm0.110$ & $0.471\pm0.098$ & $0.038\pm0.016$ & $0.767\pm0.010$ & $0.542\pm0.024$ & $0.635\pm0.024$ & $0.000\pm0.000$ & $0.658\pm0.137$ & $0.595\pm0.120$ \\
        GPT-5.6-Luna & $0.057\pm0.015$ & $0.700\pm0.005$ & $0.536\pm0.005$ & $0.666\pm0.017$ & $0.000\pm0.000$ & $0.931\pm0.054$ & $0.845\pm0.060$ & $0.033\pm0.008$ & $0.782\pm0.021$ & $0.571\pm0.044$ & $0.675\pm0.042$ & $0.005\pm0.005$ & $0.755\pm0.137$ & $0.726\pm0.129$ \\
        \hline
    \end{tabular}
    }
    \par\vspace{1pt}
    \textit{(e) Description, relational, and molecular-weight calculation}\par\smallskip
    \scriptsize
    \setlength{\tabcolsep}{3pt}
    \resizebox{0.99\textwidth}{!}{
    \begin{tabular}{|l|cccc|ccc|cccc|}
        \hline
        \multirow{2}{*}{Model}
        & \multicolumn{4}{c|}{Description Accuracy ($n=100$)}
        & \multicolumn{3}{c|}{Relational Accuracy ($n=100$)}
        & \multicolumn{4}{c|}{Molecular-Weight Calculation ($n=100$)} \\
        \cline{2-12}
        & Carbon & Ring & Functional Group & Bond Type
        & Rep. Equiv. & Chem. Sim. & Weight Comp.
        & Log RMSE & MAE & Relative RMSE & RMSE \\
        \hline \hline
        ChemDFM & $0.006\pm0.009$ & $0.001\pm0.003$ & $0.721\pm0.081$ & $0.689\pm0.106$ & $0.806\pm0.097$ & $0.475\pm0.026$ & $0.510\pm0.026$ & $0.428\pm0.064$ & $336.977\pm71.569$ & $3.404\pm3.993$ & $571.395\pm343.928$ \\
        ChemDFM-R & $0.181\pm0.069$ & $0.165\pm0.057$ & $0.811\pm0.042$ & $0.753\pm0.045$ & $0.996\pm0.005$ & $0.511\pm0.059$ & $0.680\pm0.069$ & $0.031\pm0.004$ & $20.238\pm3.294$ & $0.066\pm0.007$ & $43.367\pm2.186$ \\
        ChemDFM-v2 & $0.101\pm0.029$ & $0.129\pm0.053$ & $0.790\pm0.028$ & $0.797\pm0.040$ & $0.991\pm0.013$ & $0.444\pm0.027$ & $0.709\pm0.057$ & $0.036\pm0.008$ & $25.582\pm6.025$ & $0.075\pm0.014$ & $58.252\pm14.859$ \\
        ChemLLM & $0.000\pm0.000$ & $0.000\pm0.000$ & $0.641\pm0.072$ & $0.647\pm0.088$ & $0.690\pm0.125$ & $0.434\pm0.045$ & $0.499\pm0.020$ & $0.150\pm0.017$ & $97.953\pm13.309$ & $0.304\pm0.044$ & $158.860\pm14.200$ \\
        ChemM-R1 & $0.068\pm0.019$ & $0.079\pm0.056$ & $0.639\pm0.068$ & $0.537\pm0.045$ & $0.560\pm0.151$ & $0.444\pm0.043$ & $0.499\pm0.013$ & $7.117\pm10.233$ & $2.06\!\times\!10^{124}\pm5.44\!\times\!10^{124}$ & $6.27\!\times\!10^{122}\pm1.66\!\times\!10^{123}$ & $2.06\!\times\!10^{125}\pm5.44\!\times\!10^{125}$ \\
        ChemR & $0.040\pm0.017$ & $0.236\pm0.055$ & $0.751\pm0.046$ & $0.657\pm0.102$ & $0.655\pm0.091$ & $0.522\pm0.029$ & $0.516\pm0.029$ & $1.501\pm1.139$ & $2.03\!\times\!10^{30}\pm5.37\!\times\!10^{30}$ & $3.99\!\times\!10^{28}\pm1.05\!\times\!10^{29}$ & $2.03\!\times\!10^{31}\pm5.37\!\times\!10^{31}$ \\
        ChemVLM-8B & $0.000\pm0.000$ & $0.000\pm0.000$ & $0.684\pm0.029$ & $0.695\pm0.069$ & $0.945\pm0.036$ & $0.419\pm0.003$ & $0.492\pm0.032$ & $0.192\pm0.111$ & $47.742\pm2.910$ & $0.158\pm0.015$ & $127.838\pm16.456$ \\
        DeepSeek-R1-Direct & $0.005\pm0.007$ & $0.109\pm0.011$ & $0.505\pm0.009$ & $0.542\pm0.112$ & $0.490\pm0.000$ & $0.580\pm0.000$ & $0.510\pm0.000$ & $1.405\pm0.558$ & $477.070\pm116.569$ & $2.361\pm1.160$ & $559.097\pm128.864$ \\
        Falcon & $0.007\pm0.008$ & $0.198\pm0.054$ & $0.500\pm0.000$ & $0.500\pm0.000$ & $0.490\pm0.000$ & $0.573\pm0.007$ & $0.511\pm0.003$ & $2.320\pm0.266$ & $1.25\!\times\!10^{15}\pm3.31\!\times\!10^{15}$ & $2.46\!\times\!10^{13}\pm6.50\!\times\!10^{13}$ & $1.25\!\times\!10^{16}\pm3.31\!\times\!10^{16}$ \\
        Galactica & $0.005\pm0.007$ & $0.058\pm0.055$ & $0.500\pm0.000$ & $0.500\pm0.000$ & $0.490\pm0.000$ & $0.580\pm0.000$ & $0.510\pm0.000$ & $0.491\pm0.079$ & $246.969\pm10.159$ & $0.664\pm0.025$ & $300.211\pm5.947$ \\
        Gemma2 & $0.054\pm0.038$ & $0.115\pm0.019$ & $0.500\pm0.000$ & $0.500\pm0.000$ & $0.512\pm0.007$ & $0.539\pm0.060$ & $0.499\pm0.018$ & $0.383\pm0.035$ & $341.856\pm21.724$ & $1.036\pm0.093$ & $412.896\pm22.762$ \\
        Gemma3 & $0.000\pm0.000$ & $0.000\pm0.000$ & $0.711\pm0.111$ & $0.740\pm0.082$ & $0.865\pm0.160$ & $0.507\pm0.062$ & $0.508\pm0.016$ & $0.097\pm0.025$ & $63.163\pm19.622$ & $0.275\pm0.083$ & $93.325\pm24.737$ \\
        General-Reasoner & $0.010\pm0.007$ & $0.216\pm0.056$ & $0.477\pm0.025$ & $0.477\pm0.024$ & $0.485\pm0.007$ & $0.451\pm0.104$ & $0.497\pm0.027$ & $1.856\pm0.101$ & $345.701\pm2.641$ & $0.964\pm0.005$ & $380.578\pm4.238$ \\
        Kale-Chem & $0.065\pm0.072$ & $0.198\pm0.027$ & $0.734\pm0.042$ & $0.819\pm0.059$ & $0.629\pm0.172$ & $0.570\pm0.065$ & $0.525\pm0.038$ & $0.165\pm0.012$ & $98.492\pm10.436$ & $0.281\pm0.026$ & $159.984\pm12.312$ \\
        Llama2 & $0.005\pm0.007$ & $0.011\pm0.012$ & $0.500\pm0.000$ & $0.497\pm0.007$ & $0.490\pm0.000$ & $0.496\pm0.068$ & $0.494\pm0.009$ & $0.420\pm0.088$ & $244.542\pm29.544$ & $0.684\pm0.092$ & $304.783\pm28.671$ \\
        Llama3 & $0.088\pm0.060$ & $0.120\pm0.086$ & $0.756\pm0.070$ & $0.663\pm0.084$ & $0.586\pm0.081$ & $0.549\pm0.059$ & $0.506\pm0.011$ & $1.348\pm1.080$ & $7.55\!\times\!10^{23}\pm1.99\!\times\!10^{24}$ & $7.84\!\times\!10^{22}\pm2.07\!\times\!10^{23}$ & $7.54\!\times\!10^{24}\pm1.99\!\times\!10^{25}$ \\
        Mistral & $0.037\pm0.021$ & $0.126\pm0.089$ & $0.709\pm0.071$ & $0.641\pm0.105$ & $0.900\pm0.122$ & $0.547\pm0.045$ & $0.520\pm0.019$ & $0.284\pm0.066$ & $155.874\pm49.624$ & $0.517\pm0.113$ & $218.228\pm45.190$ \\
        Qwen2 & $0.043\pm0.046$ & $0.246\pm0.127$ & $0.670\pm0.075$ & $0.645\pm0.089$ & $0.883\pm0.063$ & $0.420\pm0.000$ & $0.581\pm0.048$ & $0.152\pm0.061$ & $86.980\pm25.650$ & $0.281\pm0.062$ & $132.880\pm30.379$ \\
        Qwen3 & $0.079\pm0.023$ & $0.220\pm0.077$ & $0.855\pm0.057$ & $0.863\pm0.063$ & $0.986\pm0.005$ & $0.571\pm0.049$ & $0.515\pm0.046$ & $0.126\pm0.034$ & $75.987\pm20.862$ & $0.269\pm0.084$ & $140.187\pm22.852$ \\
        RetroDFM-R & $0.003\pm0.004$ & $0.015\pm0.012$ & $0.523\pm0.027$ & $0.501\pm0.012$ & $0.446\pm0.061$ & $0.579\pm0.003$ & $0.509\pm0.003$ & $6.517\pm6.139$ & $1.99\!\times\!10^{172}\pm5.27\!\times\!10^{172}$ & $4.17\!\times\!10^{170}\pm1.10\!\times\!10^{171}$ & $1.96\!\times\!10^{173}\pm5.19\!\times\!10^{173}$ \\
        \hline
        Claude-Sonnet-5 & $0.130\pm0.055$ & $0.682\pm0.087$ & $0.826\pm0.009$ & $0.959\pm0.009$ & $1.000\pm0.000$ & $0.569\pm0.038$ & $0.701\pm0.043$ & $2.156\pm0.183$ & $348.283\pm6.173$ & $0.968\pm0.016$ & $387.573\pm2.523$ \\
        Gemini-3.5-Flash-Lite & $0.050\pm0.019$ & $0.414\pm0.052$ & $0.525\pm0.026$ & $0.514\pm0.016$ & $0.610\pm0.066$ & $0.470\pm0.085$ & $0.460\pm0.061$ & $1.767\pm0.095$ & $327.779\pm18.430$ & $1.044\pm0.317$ & $394.916\pm73.859$ \\
        GPT-5.6-Luna & $0.324\pm0.101$ & $0.657\pm0.065$ & $0.881\pm0.024$ & $0.925\pm0.048$ & $1.000\pm0.000$ & $0.633\pm0.050$ & $0.787\pm0.052$ & $0.049\pm0.012$ & $34.952\pm9.247$ & $0.123\pm0.032$ & $53.390\pm9.070$ \\
        \hline
    \end{tabular}
    }
    \end{minipage}
    \end{adjustbox}
\end{table*}

\begin{table*}[t]
    \centering
    \caption{Image-task results for vision-capable models (mean $\pm$ SD over evaluated conditions).}
    \label{tab:image_task_results}
    \tiny
    \setlength{\tabcolsep}{3.5pt}
    \resizebox{0.98\textwidth}{!}{
    \begin{tabular}{|l|cccc|cccc|}
        \hline
        \multirow{2}{*}{Model}
        & \multicolumn{4}{c|}{Image $\rightarrow$ SMILES ($n=100$)}
        & \multicolumn{4}{c|}{Image $\rightarrow$ IUPAC ($n=100$)} \\
        \cline{2-9}
        & Exact Match & MACCS & Morgan & RDK & Exact Match & MACCS & Morgan & RDK \\
        \hline \hline
        ChemM-R1 & $0.195\pm0.021$ & $0.871\pm0.037$ & $0.679\pm0.045$ & $0.717\pm0.055$ & $0.233\pm0.205$ & $0.787\pm0.066$ & $0.515\pm0.165$ & $0.665\pm0.086$ \\
        ChemVLM-8B & $0.200\pm0.012$ & $0.869\pm0.004$ & $0.634\pm0.008$ & $0.690\pm0.008$ & $0.006\pm0.010$ & $0.587\pm0.013$ & $0.231\pm0.009$ & $0.324\pm0.010$ \\
        Gemma3 & $0.000\pm0.000$ & $0.375\pm0.016$ & $0.124\pm0.007$ & $0.219\pm0.015$ & $0.000\pm0.000$ & $0.525\pm0.011$ & $0.192\pm0.003$ & $0.292\pm0.012$ \\
        \hline
        Claude-Sonnet-5 & $0.633\pm0.028$ & $0.981\pm0.005$ & $0.902\pm0.016$ & $0.935\pm0.013$ & $0.687\pm0.039$ & $0.980\pm0.002$ & $0.903\pm0.014$ & $0.922\pm0.011$ \\
        Gemini-3.5-Flash-Lite & $0.300\pm0.031$ & $0.915\pm0.007$ & $0.746\pm0.015$ & $0.775\pm0.010$ & $0.095\pm0.024$ & $0.830\pm0.011$ & $0.514\pm0.017$ & $0.553\pm0.015$ \\
        GPT-5.6-Luna & $0.122\pm0.041$ & $0.875\pm0.008$ & $0.584\pm0.017$ & $0.654\pm0.014$ & $0.124\pm0.014$ & $0.833\pm0.008$ & $0.500\pm0.008$ & $0.560\pm0.009$ \\
        \hline
    \end{tabular}
    }

    \vspace{6pt}

    \resizebox{0.98\textwidth}{!}{
    \begin{tabular}{|l|cccc|cccc|}
        \hline
        \multirow{2}{*}{Model}
        & \multicolumn{4}{c|}{Forward Prediction (Image) ($n=100$)}
        & \multicolumn{4}{c|}{Retrosynthesis (Image) ($n=100$)} \\
        \cline{2-9}
        & Exact Match & MACCS & Morgan & RDK & Exact Match & MACCS & Morgan & RDK \\
        \hline \hline
        ChemM-R1 & $0.000\pm0.000$ & $0.613\pm0.032$ & $0.342\pm0.019$ & $0.423\pm0.034$ & $0.000\pm0.000$ & $0.744\pm0.011$ & $0.495\pm0.010$ & $0.583\pm0.015$ \\
        ChemVLM-8B & $0.000\pm0.000$ & $0.607\pm0.009$ & $0.327\pm0.008$ & $0.444\pm0.007$ & $0.000\pm0.000$ & $0.735\pm0.003$ & $0.451\pm0.005$ & $0.555\pm0.006$ \\
        Gemma3 & $0.000\pm0.000$ & $0.334\pm0.027$ & $0.118\pm0.001$ & $0.199\pm0.005$ & $0.000\pm0.000$ & $0.366\pm0.015$ & $0.131\pm0.004$ & $0.218\pm0.020$ \\
        \hline
        Claude-Sonnet-5 & $0.112\pm0.008$ & $0.739\pm0.007$ & $0.555\pm0.008$ & $0.631\pm0.010$ & $0.113\pm0.026$ & $0.677\pm0.097$ & $0.514\pm0.075$ & $0.587\pm0.091$ \\
        Gemini-3.5-Flash-Lite & $0.028\pm0.008$ & $0.630\pm0.017$ & $0.423\pm0.013$ & $0.486\pm0.020$ & $0.003\pm0.004$ & $0.744\pm0.016$ & $0.513\pm0.012$ & $0.612\pm0.012$ \\
        GPT-5.6-Luna & $0.005\pm0.005$ & $0.626\pm0.018$ & $0.341\pm0.016$ & $0.428\pm0.022$ & $0.000\pm0.000$ & $0.733\pm0.009$ & $0.420\pm0.005$ & $0.521\pm0.006$ \\
        \hline
    \end{tabular}
    }
\end{table*}

\begin{table*}[!t]
    \centering
    \caption{Results for tasks without prompt variation: (a) design and analytical tasks, and (b) literature-grounded QA and numerical tasks.}
    \label{tab:no_template_results}
    \renewcommand{\arraystretch}{0.82}
    \begin{adjustbox}{max totalsize={0.99\textwidth}{0.84\textheight},center}
    \begin{minipage}{\textwidth}
    \centering
    \textit{(a) Description-based design, open response, and true/false}\par\smallskip
    \tiny
    \setlength{\tabcolsep}{4pt}
    \resizebox{0.98\textwidth}{!}{
    \begin{tabular}{|l|cccc|cccc|c|}
        \hline
        \multirow{2}{*}{Model}
        & \multicolumn{4}{c|}{Description-Based Design ($n=100$)}
        & \multicolumn{4}{c|}{Open Response ($n=100$)}
        & \multicolumn{1}{c|}{True/False ($n=100$)} \\
        \cline{2-10}
        & Exact Match & MACCS & Morgan & RDK & Exact Match & ROUGE-1 & ROUGE-2 & ROUGE-L & Accuracy \\
        \hline \hline
        ChemDFM & $0.500$ & $0.946$ & $0.850$ & $0.905$ & $0.020$ & $0.148$ & $0.035$ & $0.147$ & $0.610$ \\
        ChemDFM-R & $0.520$ & $0.946$ & $0.837$ & $0.892$ & $0.040$ & $0.193$ & $0.057$ & $0.184$ & $0.230$ \\
        ChemDFM-v2 & $0.440$ & $0.936$ & $0.805$ & $0.857$ & $0.030$ & $0.191$ & $0.066$ & $0.179$ & $0.270$ \\
        ChemLLM & $0.000$ & $0.622$ & $0.264$ & $0.393$ & $0.020$ & $0.086$ & $0.013$ & $0.085$ & $0.140$ \\
        ChemM-R1 & \multicolumn{4}{c|}{\textit{Unable to complete}} & $0.000$ & $0.033$ & $0.014$ & $0.030$ & $0.180$ \\
        ChemR & $0.350$ & $0.888$ & $0.746$ & $0.807$ & $0.000$ & $0.025$ & $0.012$ & $0.025$ & $0.430$ \\
        ChemVLM-8B & $0.040$ & $0.549$ & $0.326$ & $0.394$ & $0.170$ & $0.287$ & $0.094$ & $0.283$ & $0.120$ \\
        DeepSeek-R1-Direct & \multicolumn{4}{c|}{\textit{Unable to complete}} & $0.000$ & $0.064$ & $0.019$ & $0.061$ & $0.640$ \\
        Falcon & \multicolumn{4}{c|}{\textit{Unable to complete}} & $0.000$ & $0.052$ & $0.012$ & $0.051$ & $0.640$ \\
        Galactica & $0.000$ & $0.308$ & $0.168$ & $0.170$ & $0.070$ & $0.114$ & $0.021$ & $0.114$ & $0.640$ \\
        Gemma2 & $0.020$ & $0.523$ & $0.338$ & $0.354$ & $0.000$ & $0.018$ & $6.64\times10^{-4}$ & $0.018$ & $0.600$ \\
        Gemma3 & $0.000$ & $0.348$ & $0.102$ & $0.179$ & $0.120$ & $0.264$ & $0.086$ & $0.264$ & $0.270$ \\
        General-Reasoner & $0.010$ & $0.500$ & $0.268$ & $0.310$ & $0.030$ & $0.150$ & $0.059$ & $0.147$ & $0.180$ \\
        Kale-Chem & $0.090$ & $0.819$ & $0.535$ & $0.655$ & $0.200$ & $0.323$ & $0.094$ & $0.323$ & $0.500$ \\
        Llama2 & $0.000$ & $0.106$ & $0.024$ & $0.049$ & $0.000$ & $0.024$ & $0.003$ & $0.024$ & $0.630$ \\
        Llama3 & $0.030$ & $0.531$ & $0.255$ & $0.361$ & $0.000$ & $0.268$ & $0.117$ & $0.268$ & $0.250$ \\
        Mistral & $0.000$ & $0.427$ & $0.156$ & $0.218$ & $0.050$ & $0.216$ & $0.093$ & $0.215$ & $0.120$ \\
        Qwen2 & $0.000$ & $0.297$ & $0.100$ & $0.156$ & $0.120$ & $0.239$ & $0.058$ & $0.239$ & $0.170$ \\
        Qwen3 & $0.010$ & $0.503$ & $0.224$ & $0.295$ & $0.170$ & $0.344$ & $0.117$ & $0.339$ & $0.180$ \\
        RetroDFM-R & $0.000$ & $0.467$ & $0.119$ & $0.206$ & $0.000$ & $0.000$ & $0.000$ & $0.000$ & $0.010$ \\
        \hline
        Claude-Sonnet-5 & $0.280$ & $0.926$ & $0.765$ & $0.848$ & $0.170$ & $0.411$ & $0.160$ & $0.404$ & $0.510$ \\
        Gemini-3.5-Flash-Lite & $0.100$ & $0.868$ & $0.574$ & $0.709$ & $0.230$ & $0.469$ & $0.197$ & $0.463$ & $0.490$ \\
        GPT-5.6-Luna & $0.090$ & $0.843$ & $0.534$ & $0.659$ & $0.160$ & $0.414$ & $0.179$ & $0.409$ & $0.250$ \\
        \hline
    \end{tabular}
    }
    \par\vspace{2pt}
    \textit{(b) Literature-grounded QA, physical-unit calculation, and yield prediction}\par\smallskip
    \tiny
    \setlength{\tabcolsep}{4pt}
    \resizebox{0.98\textwidth}{!}{
    \begin{tabular}{|l|cccc|cccc|cccc|}
        \hline
        \multirow{2}{*}{Model}
        & \multicolumn{4}{c|}{Literature-Grounded QA ($n=211$)}
        & \multicolumn{4}{c|}{Physical-Unit Calculation ($n=392$)}
        & \multicolumn{4}{c|}{Yield Prediction ($n=300$)} \\
        \cline{2-13}
        & Exact Match & ROUGE-1 & ROUGE-2 & ROUGE-L & Log RMSE & MAE & Relative RMSE & RMSE & Log RMSE & MAE & Relative RMSE & RMSE \\
        \hline \hline
        ChemDFM & $0.019$ & $0.225$ & $0.147$ & $0.220$ & $4.197$ & $3.04\times10^{22}$ & $1.03\times10^{23}$ & $2.60\times10^{23}$ & $0.754$ & $310.604$ & $18.953$ & $1142.241$ \\
        ChemDFM-R & $0.038$ & $0.253$ & $0.163$ & $0.248$ & $3.159$ & $2.47\times10^{22}$ & $5.08\times10^{20}$ & $2.37\times10^{23}$ & $0.279$ & $23.252$ & $1.403$ & $30.730$ \\
        ChemDFM-v2 & $0.047$ & $0.273$ & $0.185$ & $0.269$ & $2.402$ & $1.71\times10^{22}$ & $5.08\times10^{20}$ & $2.08\times10^{23}$ & $0.268$ & $23.861$ & $1.395$ & $30.917$ \\
        ChemLLM & $0.043$ & $0.163$ & $0.076$ & $0.159$ & $2.707$ & $9.88\times10^{21}$ & $3.34\times10^{20}$ & $1.29\times10^{23}$ & $0.238$ & $37.066$ & $1.134$ & $46.565$ \\
        ChemM-R1 & $0.009$ & $0.104$ & $0.039$ & $0.102$ & $10.656$ & $2.89\times10^{125}$ & $5.48\times10^{125}$ & $4.54\times10^{126}$ & $2.462$ & $3.36\times10^{36}$ & $7.83\times10^{35}$ & $5.79\times10^{37}$ \\
        ChemR & $0.062$ & $0.158$ & $0.075$ & $0.157$ & $5.483$ & $6.39\times10^{28}$ & $8.53\times10^{44}$ & $1.26\times10^{30}$ & $1.674$ & $1.32\times10^{10}$ & $2.29\times10^{9}$ & $2.29\times10^{11}$ \\
        ChemVLM-8B & $0.071$ & $0.306$ & $0.210$ & $0.302$ & $2.204$ & $4.23\times10^{21}$ & $5.21\times10^{20}$ & $8.07\times10^{22}$ & $0.333$ & $34.531$ & $2.030$ & $42.913$ \\
        DeepSeek-R1-Direct & $0.005$ & $0.163$ & $0.096$ & $0.159$ & $6.052$ & $2.81\times10^{35}$ & $2.54\times10^{34}$ & $5.56\times10^{36}$ & $1.663$ & $61.993$ & $1.196$ & $67.556$ \\
        Falcon & $0.000$ & $0.172$ & $0.107$ & $0.168$ & $5.897$ & $1.11\times10^{23}$ & $6.83\times10^{22}$ & $9.45\times10^{23}$ & $1.643$ & $55.865$ & $1.113$ & $62.553$ \\
        Galactica & $0.038$ & $0.211$ & $0.114$ & $0.209$ & $17.918$ & $2.42\times10^{125}$ & $1.58\times10^{124}$ & $1.70\times10^{126}$ & $0.333$ & $34.557$ & $2.030$ & $42.921$ \\
        Gemma2 & $0.033$ & $0.093$ & $0.052$ & $0.093$ & $13.337$ & $1.28\times10^{125}$ & $2.53\times10^{127}$ & $1.13\times10^{126}$ & $0.333$ & $34.557$ & $2.030$ & $42.921$ \\
        Gemma3 & $0.137$ & $0.348$ & $0.194$ & $0.345$ & $12.952$ & $3.12\times10^{125}$ & $4.58\times10^{121}$ & $3.93\times10^{126}$ & $0.419$ & $30.932$ & $1.872$ & $39.558$ \\
        General-Reasoner & $0.014$ & $0.184$ & $0.117$ & $0.181$ & $3.732$ & $2.89\times10^{22}$ & $7.14\times10^{21}$ & $2.59\times10^{23}$ & $1.587$ & $63.098$ & $0.959$ & $68.143$ \\
        Kale-Chem & $0.133$ & $0.393$ & $0.230$ & $0.389$ & $4.072$ & $3.04\times10^{22}$ & $9.30\times10^{17}$ & $2.60\times10^{23}$ & $1.464$ & $39.591$ & $1.432$ & $49.134$ \\
        Llama2 & $0.000$ & $0.071$ & $0.041$ & $0.070$ & $22.403$ & $3.02\times10^{125}$ & $2.81\times10^{126}$ & $1.82\times10^{126}$ & $0.338$ & $42.960$ & $1.909$ & $51.793$ \\
        Llama3 & $0.066$ & $0.379$ & $0.256$ & $0.377$ & $3.914$ & $5.58\times10^{22}$ & $1.56\times10^{22}$ & $5.83\times10^{23}$ & $1.662$ & $60.129$ & $0.998$ & $66.060$ \\
        Mistral & $0.052$ & $0.245$ & $0.135$ & $0.239$ & $14.042$ & $4.17\times10^{125}$ & $2.54\times10^{126}$ & $5.03\times10^{126}$ & $0.976$ & $50.474$ & $0.782$ & $56.723$ \\
        Qwen2 & $0.152$ & $0.383$ & $0.251$ & $0.379$ & $3.426$ & $3.04\times10^{22}$ & $5.05\times10^{26}$ & $2.60\times10^{23}$ & $0.328$ & $33.293$ & $1.993$ & $41.800$ \\
        Qwen3 & $0.185$ & $0.428$ & $0.292$ & $0.424$ & $1.855$ & $4.22\times10^{23}$ & $5.08\times10^{19}$ & $8.15\times10^{24}$ & $0.288$ & $24.403$ & $1.657$ & $32.266$ \\
        RetroDFM-R & $0.000$ & $0.003$ & $0.000$ & $0.003$ & $4.069$ & $2.19\times10^{22}$ & $3.17\times10^{22}$ & $2.22\times10^{23}$ & $2.665$ & $65.073$ & $0.992$ & $69.950$ \\
        \hline
        Claude-Sonnet-5 & $0.137$ & $0.350$ & $0.218$ & $0.344$ & $3.764$ & $3.04\times10^{22}$ & $3.10\times10^{21}$ & $2.60\times10^{23}$ & $0.905$ & $41.611$ & $1.534$ & $52.305$ \\
        Gemini-3.5-Flash-Lite & $0.185$ & $0.419$ & $0.270$ & $0.415$ & $2.290$ & $8.58\times10^{21}$ & $9.56\times10^{8}$ & $1.20\times10^{23}$ & $1.134$ & $86.236$ & $4.633$ & $134.877$ \\
        GPT-5.6-Luna & $0.133$ & $0.383$ & $0.242$ & $0.374$ & $3.266$ & $3.04\times10^{22}$ & $7.08\times10^{13}$ & $2.60\times10^{23}$ & $0.277$ & $25.068$ & $1.421$ & $31.589$ \\
        \hline
    \end{tabular}
    }
    \end{minipage}
    \end{adjustbox}
\end{table*}

\subsubsection{Mechanistic}

Mechanistic tasks require reasoning over chemical transformations and reaction pathways. These tasks are defined as
\[
LLM(p_{\text{react}}) = m,
\]
where $p_{\text{react}}$ specifies the reaction context and the output is the predicted molecular structure $m$.

We include two mechanistic tasks: forward prediction, in which the major product of a reaction is predicted, and retrosynthesis, in which reactants are inferred from products and reagents. Each is evaluated with SMILES, molecular-formula, and molecular-graph-image inputs. SMILES examples are drawn from ChemLLMBench \cite{guo2023can}, formula examples from ChemAlgebra \cite{valenti2024chemalgebra}, and corresponding graph images are generated with RDKit \cite{landrum2013rdkit}. SMILES- and image-based settings are scored using exact match and MACCS, Morgan, and RDK fingerprint similarity. Because formulas lack explicit bonding information, formula-based settings instead use exact match, cosine similarity between elemental-count vectors (atom cosine), and atom-level F1. Exact match requires complete reconstruction, whereas atom cosine and F1 quantify partial compositional agreement.

\subsubsection{Relational}

Relational tasks require reasoning over relationships between two or more molecular entities. These tasks are defined as
\[
LLM(m_1, \ldots, m_n, \tau) = y^{(s)},
\]
where $\{m_i\}_{i=1}^n$ is a set of molecules, $\tau$ specifies the relational objective, and $y^{(s)}$ describes the relationship among the inputs. We include molecular representation equivalence, chemical-similarity comparison, and molecular-weight comparison. Representation equivalence determines whether two representations denote the same molecule, chemical-similarity comparison selects which candidate is more similar to a query molecule, and molecular-weight comparison selects the heavier or lighter of two molecules. Each subtask is rendered with both SMILES and IUPAC inputs, and all three are evaluated using accuracy. Relational tasks therefore embed a secondary task, namely the relational objective evaluated across the molecular inputs.

\subsection{Subset Selection}


To control evaluation cost when a source test set exceeds the evaluation budget, we construct a representative subset. Each example is encoded as a binary vector indicating which reference models answer it correctly. We cluster these vectors using $K$-means, where $K$ is the desired subset size, and initially select the example nearest each cluster centroid.

We then compare each reference model's subset accuracy with its full-test accuracy. Starting with the model with the largest deviation, we swap examples within the subset in order to reduce deviation without increasing another model's error beyond the previous maximum. This process continues until every reference model is within one percentage point of its full-test accuracy. We select the smallest subset satisfying this criterion, subject to a minimum of 100 and a maximum of 500 examples. Table \ref{tab:hiv_subset} illustrates the resulting agreement for HIV.

\subsection{Instruction Generation}



LLMs are sensitive to semantically equivalent changes in prompt wording \cite{mizrahi2024state}. In our ClinTox evaluation, prompt choice changes model accuracy by as much as 81 percentage points (Figure \ref{fig:prompt_difference}). We therefore evaluate applicable tasks under multiple templates rather than a single fixed instruction.

Following ChemBench4K \cite{zhang2024chemllm}, we use ChatGPT (GPT-5 Pro) to generate ten templates per task and manually verify that they preserve the same objective, required inputs, and expected output. To control evaluation costs while retaining instruction variation, we select the first four verified templates per task for the reported evaluation (the GitHub repository contains all ten prompts for all tasks). For description, molecular-weight calculation, and relational settings, the four templates are evaluated with both SMILES and IUPAC inputs, producing eight prompt--representation conditions. Tables \ref{tab:template_task_results} and \ref{tab:image_task_results} report the resulting means and population standard deviations. Table \ref{tab:prompts_clintox_example} presents the ten verified ClinTox templates.

Analytical questions, description-based design, physical-unit calculation, and yield prediction do not use interchangeable templates. For these tasks, we retain the source questions after first manually cleaning and standardizing each question.

\subsection{Evaluation Setup}

We evaluate 23 models zero-shot with the same system prompt and task-specific output-format instructions across models. Image settings include only the six vision-capable models. Outputs are restricted by task to real-valued numbers, coefficient vectors, binary labels, molecular names, SMILES strings, molecular formulas, or short answers. Direct-answer models have a 128-token limit. Models with explicit reasoning traces have 256 tokens to accommodate thinking. We apply common answer extraction and normalization before scoring. ``Unable to complete'' indicates that all responses are missing, unusable, or otherwise invalid.

\section{Results}

Tables \ref{tab:template_task_results}--\ref{tab:no_template_results} report the baseline results on ChemDIRT. Reported standard deviations measure variation across prompts and, in panel (e) of Table \ref{tab:template_task_results}, molecular representations. Note that they are not confidence intervals over examples. Most prompt-varied settings use four prompts, whereas panel (e) aggregates eight conditions (four prompts $\times$ SMILES and IUPAC inputs). Entries marked ``Unable to complete'' indicate that the model returned no usable responses. The central finding is not a single overall winner, but substantial variation both across tasks and across prompt formulations. Every model exhibits at least one major weakness: models that are strong on molecular translation or reaction prediction can be poor on regression, literature-grounded or open-response QA, or general chemistry reasoning, while models that perform well on language-oriented tasks often trail specialized chemistry models on structure generation. Thus, performance on any one task family is a poor proxy for general chemical reasoning ability.

Figure \ref{fig:template_spread} makes the within-task instability explicit. On ClinTox, Llama3 ranges from 0.095 to 0.905 accuracy across semantically equivalent prompts. Falcon ranges from 0.095 to 0.824, and GPT-5.6-Luna from 0.236 to 0.899. The same pattern extends beyond classification: Falcon's coefficient-balancing accuracy spans 0.400--0.910, while Galactica's atom-cosine score on formula-based retrosynthesis spans 0.100--0.991. These gaps are large enough to reverse apparent model rankings depending on the selected template. The mean and standard deviation in Table \ref{tab:template_task_results} therefore provide a more representative summary than any single-prompt score.

The strongest systems are also highly specialized. ChemDFM-R and ChemDFM-v2 are consistently strong on name translation, molecular classification, and mechanistic tasks, yet their open-response exact match is at most 0.040 and their true/false accuracy is at most 0.270. RetroDFM-R achieves the highest SMILES-retrosynthesis exact match (0.685), but remains unable to complete SMILES-to-IUPAC translation, reaches only 0.150 accuracy on ClinTox, and produces extreme errors on lipophilicity, molecular-weight, and physical-unit calculation. Claude-Sonnet-5 dominates the image-to-structure tasks, reaching 0.633 exact match for image-to-SMILES and 0.687 for image-to-IUPAC, but its formula-based mechanistic scores are substantially lower than those of the strongest chemistry-specialized models. Conversely, models such as Falcon, Llama2, and DeepSeek-R1-Direct obtain zero exact match on most structure-generation tasks and remain weak across many other task families despite isolated successes.

Numerical tasks expose an additional failure mode. Although some models attain comparatively low errors on ESOL, lipophilicity, or FreeSolv, no model performs reliably across all regression and computation settings. ChemM-R1, ChemR, Gemma3, RetroDFM-R, and several general-purpose baselines produce errors many orders of magnitude larger than their peers on at least one task. Oxygen-permeability prediction is especially difficult as most models either cluster around a high error floor or generate extreme outliers, and performance varies sharply across templates. Physical-unit calculation shows a similarly broad failure pattern, whereas yield prediction is comparatively tractable for a subset of models. These results indicate that a model's ability to produce fluent chemical text or valid molecular strings does not imply reliable numerical reasoning.

Representation changes further alter the model ordering. Figure \ref{fig:representation_gap} shows that IUPAC has gains of 0.273 over SMILES for Kale-Chem on representation equivalence and 0.163 for Mistral on ring count, but a 0.133 SMILES advantage for Llama2 on chemical similarity. Direction also matters for name translation as success on IUPAC-to-SMILES does not guarantee comparable SMILES-to-IUPAC performance. The same asymmetry appears in reaction prediction, where performance differs between forward prediction and retrosynthesis for SMILES, molecular-formula, and image inputs. For example, ChemR is among the strongest models on SMILES-based reactions but falls well behind the leading systems on formula-based reactions, while RetroDFM-R shows the opposite specialization between the forward and retrosynthetic SMILES settings. Taken together, Tables \ref{tab:template_task_results}--\ref{tab:no_template_results} show that current models are neither prompt-invariant nor broadly capable across ChemDIRT. Strong performance on narrow evaluation settings can conceal severe task-specific, representation-specific, and instruction-specific weaknesses.

\section{Discussion}

ChemDIRT shows that current LLMs, including chemistry-specialized systems, do not exhibit robust general chemical reasoning. Many succeed only within narrow task or representation families, and performance can change sharply under equivalent prompt wording. These results caution against broad capability claims based on a single task or template. They also frame evaluation diversity as a benchmark data-quality requirement: a fixed prompt or representation can introduce substantial measurement bias. Releasing the examples, verified templates, and scoring code makes these design choices auditable and supports reproducible scientific-model comparison. Tool augmentation, explicit molecular representations, constrained numerical execution, and training across equivalent instructions are promising directions for improving robustness.

\subsection{Limitations}

ChemDIRT samples representative molecular-level tasks rather than every chemistry domain. It does not yet cover laboratory execution, spectroscopy and instrument interpretation, materials synthesis planning, or long-horizon experimental workflows. Image evaluation is limited to molecular graphs and does not include laboratory photographs, spectra, or scientific figures. Prompt robustness is measured using a finite set of manually verified templates, and subset-based evaluation cannot preserve every property of the complete source datasets.

\subsection{Conclusion}

ChemDIRT evaluates LLMs across diverse chemistry tasks while reducing dependence on specific wording and molecular representations. Across 23 models, performance is uneven and often highly sensitive to the prompt, task, or representation. These findings show that current models do not yet demonstrate reliable general chemical reasoning under zero-shot evaluation.

\bibliographystyle{IEEEtran}
\bibliography{ref.bib}

\end{document}